%% file: paper-supplement-arxiv.tex
\pdfoutput=1 %
\documentclass[10pt,twocolumn,letterpaper]{article}

\usepackage[pagenumbers]{cvpr}

\input{preamble}

\definecolor{cvprblue}{rgb}{0.21,0.49,0.74}
\usepackage[pagebackref,breaklinks,colorlinks,allcolors=cvprblue]{hyperref}
\usepackage{orcidlink}

\hypersetup{
  pdftitle    = {epsilon-SegFormer: Multi-Resolution Vision Transformers for Sparsely Supervised Segmentation in Electron Microscopy},
  pdfauthor   = {Erfan Amidi, Sheida RahnamaiKordasiabi, Edoardo Giacomello, Florian Jug},
  pdfsubject  = {Semantic segmentation of electron microscopy images with sparse pixel annotations},
  pdfkeywords = {semantic segmentation, electron microscopy, sparse supervision, vision transformer, multi-resolution, masked autoencoding},
}

\def\confName{CVPR}
\def\confYear{2027}

\newcommand{\suppref}[1]{\ref{#1}}
\newcommand{\mainref}[1]{\ref{#1}}

\makeatletter
\newdimen\tbextra \newdimen\tbgapa \newdimen\tbgapb \newdimen\tbgapc \newbox\tbboxa \newbox\tbboxb \newcount\tbnum \newcount\tbtot
\def\@maketitle{%
   \newpage
   \null
   \setbox\tbboxa\vbox{\large\lineskip .5em\begin{tabular}[t]{c}Name\end{tabular}}%
   \setbox\tbboxb\vbox{\large\lineskip .5em\begin{tabular}[t]{c}\@author\end{tabular}}%
   \tbextra=\dimexpr\ht\tbboxb+\dp\tbboxb-\ht\tbboxa-\dp\tbboxa\relax
   \ifdim\tbextra<0pt \tbextra=0pt \fi
   \ifdim\tbextra>34pt \tbextra=34pt \fi %
   \tbnum=\tbextra \tbgapa=68pt \tbtot=\tbgapa
   \tbgapa=\dimexpr .375in-.375in*\tbnum/\tbtot\relax
   \tbgapb=\dimexpr 24pt-24pt*\tbnum/\tbtot\relax
   \tbgapc=\dimexpr 12pt+.5em-12pt*\tbnum/\tbtot-.5em*\tbnum/\tbtot\relax
   \vskip\tbgapa
   \begin{center}
      {\Large \bf \@title \par}
      \vspace*{\tbgapb}{%
        \large
        \lineskip .5em
        \begin{tabular}[t]{c}
          \@author
        \end{tabular}
        \par
      }
      \vspace*{\tbgapc}
   \end{center}
   \typeout{TITLEBLOCK: author block surplus over a one-line block = \the\tbextra}%
}
\makeatother



\newcommand{\tabstylesparse}{\tabstylewide}
\newcommand{\panelrow}[2]{\multicolumn{#1}{l}{\textbf{#2}}\\}
\newcommand{\dn}[1]{$-#1$}
\newcommand{\msd}[2]{#1\,{\scriptsize$\pm$#2}}
\newcommand{\fitwidth}[3]{\sbox0{\setlength{\tabcolsep}{0pt}#3}\setlength{\tabcolsep}{\dimexpr(#1-\wd0)/#2\relax}#3}

\makeatletter %
\newcommand{\suppsetup}{%
  \setcounter{topnumber}{3}\setcounter{dbltopnumber}{1}\setcounter{totalnumber}{6}
  \renewcommand{\topfraction}{0.9}\renewcommand{\dbltopfraction}{0.85}\renewcommand{\floatpagefraction}{0.95}\renewcommand{\dblfloatpagefraction}{0.5}
  \setlength{\@fptop}{0pt}\setlength{\@fpsep}{10pt plus 1fil}\setlength{\@fpbot}{0pt plus 1fil}
  \setlength{\@dblfptop}{0pt}\setlength{\@dblfpsep}{8pt plus 1fil}\setlength{\@dblfpbot}{0pt plus 1fil}
  \raggedbottom
  \renewcommand{\tabstylenarrow}{\footnotesize\setlength{\tabcolsep}{2pt}\renewcommand{\arraystretch}{1.1}}
  \renewcommand{\tabstyledbl}{\footnotesize\setlength{\tabcolsep}{4pt}\renewcommand{\arraystretch}{1.1}}
  \renewcommand{\emphrow}{\rowcolor{emphgray}}
  \renewcommand{\grouprow}[2]{\multicolumn{##1}{l}{\textbf{##2}}\\}
  \renewcommand{\sd}[1]{{\scriptsize$\pm$##1}}
  \renewenvironment{tablewide}[1][t]{\begin{table*}[##1]}{\end{table*}}
  \WarningFilter{caption}{The type was already set}
  \setcounter{table}{0}\renewcommand{\thetable}{S\arabic{table}}\renewcommand{\theHtable}{S\arabic{table}}
  \captionsetup[table]{name=Supplementary Table}
  \crefname{table}{Supp.~Tab.}{Supp.~Tabs.}\Crefname{table}{Supp.~Tab.}{Supp.~Tabs.}
  \setcounter{figure}{0}\renewcommand{\thefigure}{S\arabic{figure}}\renewcommand{\theHfigure}{S\arabic{figure}}
  \captionsetup[figure]{name=Supplementary Figure}
  \crefname{figure}{Supp.~Fig.}{Supp.~Figs.}\Crefname{figure}{Supp.~Fig.}{Supp.~Figs.}
  \crefname{appendix}{Supp.~Section}{Supp.~Sections}\Crefname{appendix}{Supp.~Section}{Supp.~Sections}
}
\makeatother

\title{\epssegformer: Multi-Resolution Vision Transformers for\\Sparsely Supervised Segmentation in Electron Microscopy}
\newcommand{\orcidErfan}{0009-0001-0598-8121}
\author{%
  Erfan Amidi\textsuperscript{1}\hspace{0.1em}\orcidlink{\orcidErfan}\hspace{1.2em}%
  Sheida RahnamaiKordasiabi\textsuperscript{2,1}\hspace{1.2em}%
  Edoardo Giacomello\textsuperscript{1}\hspace{1.2em}%
  Florian Jug\textsuperscript{1}\\[0.7em]
  {\normalsize\textsuperscript{1}Human Technopole\hspace{2em}\textsuperscript{2}Technische Universit\"at Dresden}%
}

\begin{document}
\maketitle

\input{sec/0_abstract}
\input{sec/1_intro}
\input{sec/2_related}
\input{sec/3_method}
\input{sec/4_experiments}
\input{sec/5_experiments}

\input{sec/5_results}
\input{sec/6_conclusion}

\input{sec/6_llm_statement}

{
    \small
    \bibliographystyle{ieeenat_fullname}
    \bibliography{main}
}

\clearpage
\maketitlesupplementary
\appendix
\suppsetup
\input{sec/7_appendix}

\end{document}

%% file: preamble.tex
\usepackage{cuted} %

\usepackage{currfile} %

\usepackage{booktabs} %
\usepackage{multirow} %
\usepackage{makecell} %
\usepackage{colortbl} %
\usepackage{float} %
\usepackage{placeins} %
\newcommand{\tabstyle}{\footnotesize\setlength{\tabcolsep}{3.5pt}\renewcommand{\arraystretch}{1.1}} %
\newcommand{\tabstylewide}{\small\setlength{\tabcolsep}{4pt}\renewcommand{\arraystretch}{1.1}} %
\newcommand{\tabstyledbl}{\tabstyle\renewcommand{\arraystretch}{1.0}} %
\newcommand{\tabstylenarrow}{\tabstyle\setlength{\tabcolsep}{2pt}} %
\definecolor{emphgray}{gray}{0.90} %
\newcommand{\emphrow}{\rowcolor{emphgray}} %
\newcommand{\grouprow}[2]{\multicolumn{#1}{@{}l}{\textbf{#2}}\\} %
\newcommand{\sd}[1]{{\scriptsize$\pm$#1}} %
\newenvironment{tablewide}[1][t]{\begin{figure*}[#1]\captionsetup{type*=table}}{\end{figure*}}

\usepackage{algorithm}
\usepackage{algorithmic}

\usepackage{qualgrid}

\usepackage{caption} %

\usepackage[protrusion=false]{microtype} %

\newcommand{\epssegformer}{\texorpdfstring{\mbox{$\varepsilon$-SegFormer}}{ε-SegFormer}\xspace}
\newcommand{\epsseg}{\texorpdfstring{\mbox{$\varepsilon$-Seg}}{ε-Seg}\xspace}

\newcommand{\muvitenc}{\texorpdfstring{$\text{MuViT}_{\text{ENC}}$}{MuViT-ENC}\xspace}
\newcommand{\muvit}{MuViT\xspace}

\renewcommand{\topfraction}{0.95}

\makeatletter
\renewcommand\paragraph{\@startsection{paragraph}{4}{\z@}{4pt plus 1pt minus 1pt}{-1em}{\normalfont\normalsize\bfseries}}
\makeatother
\AtEndPreamble{%
  \crefname{figure}{Figure}{Figures}\Crefname{figure}{Figure}{Figures}%
  \crefname{table}{Table}{Tables}\Crefname{table}{Table}{Tables}%
  \crefname{section}{Section}{Sections}\Crefname{section}{Section}{Sections}%
}

%% file: sec/0_abstract.tex
\begin{abstract}
Segmenting organelles in electron microscopy (EM) volumes requires dense annotations that take experts months to produce, and every gain in imaging resolution adds to this cost. 
Sparsely supervised methods reduce this cost to a small fraction of labeled pixels, but the convolutional networks they build on see a limited neighborhood around each pixel, while the identity of an organelle often depends on context far beyond it.
We introduce \epssegformer, a multi-resolution Vision Transformer for sparsely supervised semantic segmentation, which builds on the architecture of \muvit and extends the training method of \epsseg.
  \setcounter{dbltopnumber}{0}%
  \input{fig/figure1}  \input{fig/figure2}  \setcounter{dbltopnumber}{1} %
For each labeled pixel, \epssegformer samples views at three physical resolutions centered on it, embeds their patches into a single token sequence, and encodes every token's position in image coordinates, so that attention relates all views in one shared frame. 
A segmentation head classifies the pixel from the tokens around it; during training, the pixel's own token is hidden, so the label must be predicted from context rather than local appearance. 
The model is trained in a single stage, without pretraining, with an auxiliary reconstruction loss. 
On the BetaSeg FIB-SEM dataset, \epssegformer learns from about 10{,}000 labeled pixels among the 670 million voxels inside three cells, and, given the cell mask, segments nuclei, mitochondria and insulin granules in a slice of a fourth, unseen cell with an average Dice of 0.89 (0.91 after post-processing). 
Ablations show that coarser views improve accuracy and that a compact head on local tokens works best, with the wider context entering through attention in the encoder.
\end{abstract}

%% file: fig/figure1.tex
\begin{figure}[t]
  \centering
  \includegraphics[width=\linewidth,height=0.5\textheight,keepaspectratio]{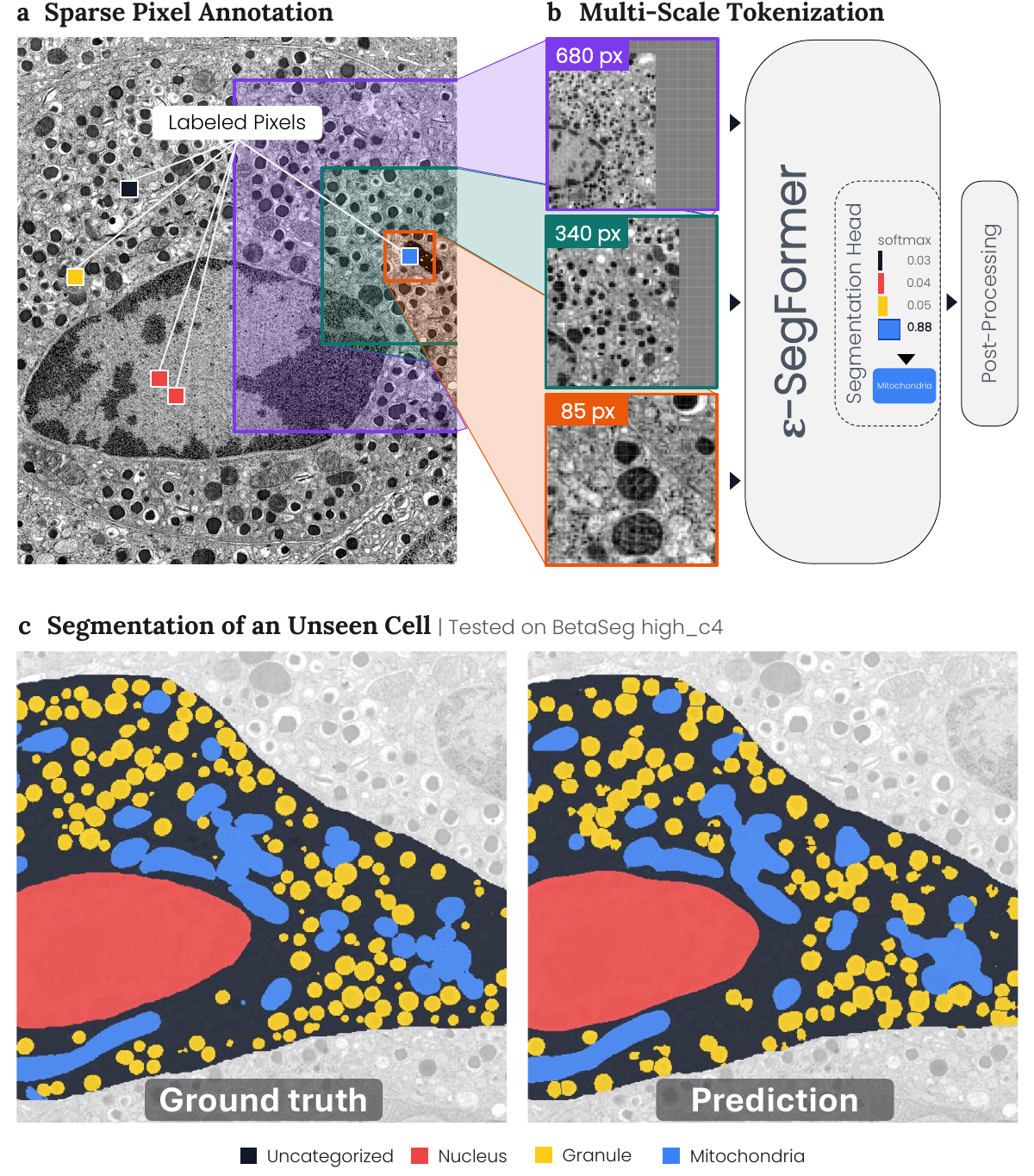}
  \caption{\textbf{Overview of \epssegformer{}.}
  \textbf{(a)}~Sparse pixel annotation: 10{,}029 labeled pixels across 2{,}611 training slices (BetaSeg c1--c3).
  \textbf{(b)}~Three views (85, 340 and 680\,px wide) centered on each labeled pixel are tokenized (867 tokens) for a ViT encoder. A segmentation head classifies the center pixel; post-processing is optional.
  \textbf{(c)}~Post-processed segmentation of an unseen slice (BetaSeg high\_c4): ground truth (left) and prediction (right).
  Classes: uncategorized (black), nucleus (red), granule (yellow), mitochondria (blue).}
  \label{fig:qualitative}
\end{figure}

%% file: fig/figure2.tex
\begin{figure*}[t]
    \centering
    \includegraphics[width=0.95\textwidth]{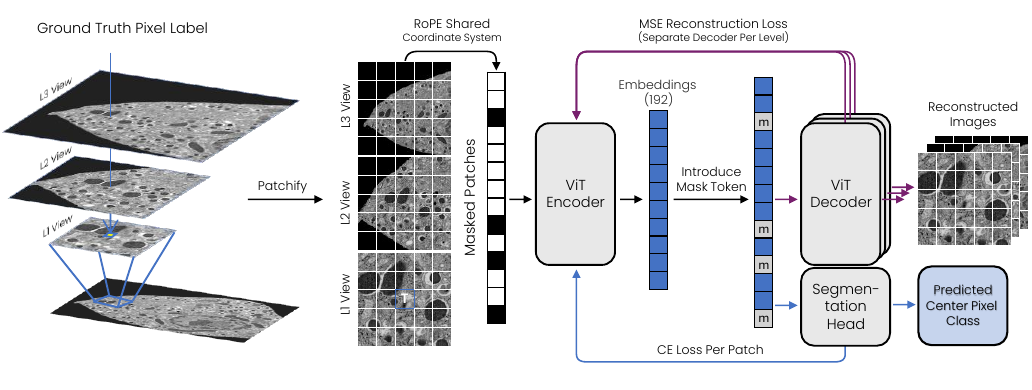}
    \caption{\textbf{Overview of \epssegformer{} Architecture.} Three pixel-centered views
    (L1: finest/highest-resolution; L2, L3: wider/coarser) are patched and combined into a single token sequence.
    2D RoPE provides a shared coordinate system.
    The encoder processes only visible tokens, dropping the center token of every view and a random subset of others. After the encoder, a learned mask token ($m$) is inserted at every dropped position. One decoder per view reconstructs the dropped patches of its view (MSE loss). A segmentation head then predicts the center pixel's class using encoded L1 tokens surrounding the center (CE loss).}
    \label{fig:muvit}
\end{figure*}

%% file: sec/1_intro.tex
\section{Introduction}
\label{sec:intro}

Volume electron microscopy (vEM) images entire cells at nanometer resolution and shows every organelle in its native context~\cite{peddie2022vem,xu2021atlas}.
Turning such volumes into biological insight requires segmenting the organelles, and a single cell imaged by focused ion beam scanning electron microscopy (FIB-SEM) easily spans hundreds of millions of voxels.
Deep networks segment organelles well when trained on dense annotations~\cite{heinrich2021cosem,conrad2023mitonet}, but producing these annotations takes experts months~\cite{conrad2021cem500k}, and models trained on one dataset often need new annotations for the next~\cite{conrad2023mitonet}.
Annotation, more than imaging, has become the bottleneck of quantitative vEM.

Sparse supervision addresses this bottleneck directly: instead of painting masks, annotators label a few scattered pixels.
Interactive tools have long trained pixel classifiers from such input~\cite{berg2019ilastik,arzt2022labkit}, point and scribble labels are established for natural images~\cite{bearman2016point,lin2016scribblesup}, and \epsseg~\cite{epsseg2025} showed that deep networks can learn to segment organelles in EM from labels on 0.05\% of the pixels or fewer.
A single labeled pixel, however, contains no information about its spatial context, although the class of a pixel is often determined by context spanning many pixels, particularly in high-resolution EM images.
A small cross-section of a mitochondrion can resemble an insulin granule, and the interior of a nucleus can look like cytoplasm until its envelope comes into view.
A model that learns from sparse labels should therefore see well beyond each labeled pixel, yet the effective receptive field of convolutional networks grows only slowly with depth~\cite{luo2016erf}, and enlarging the input at full resolution is expensive, since memory grows quadratically with the field of view in 2D; $\mu$Split therefore fed the wider surroundings at reduced resolution~\cite{ashesh2023usplit}.

We address this with \epssegformer, which brings the multi-resolution Vision Transformer of MuViT~\cite{muvit2026} to learning from sparse labels.
For each labeled pixel, we sample three views centered on it, with increasing field of view and decreasing resolution, split all of them into tokens, and encode the joint sequence with one transformer whose rotary position embeddings~\cite{rope,heo2024ropevit} place all views in the same slice coordinates.
With 867 tokens, the model sees 680 pixels (about 11\,$\mu$m) across while keeping full resolution near the labeled pixel (\cref{fig:qualitative,fig:muvit}).
Each coarser view widens the field of view at the same token cost, so the cost of context grows with the number of views rather than with the area covered; the price is that distant context is seen at lower resolution.

Learning from single labeled pixels also invites a shortcut~\cite{geirhos2020shortcut}: the token that contains the labeled pixel suffices to fit the labels from local appearance.
Extending the center inpainting of \epsseg and following the blind-spot principle of Noise2Void~\cite{krull2019noise2void}, we drop this token from every view during training, so that the segmentation head must classify the pixel from its context, while an auxiliary loss reconstructs the dropped patches, as in MAE~\cite{mae}.
Unlike MAE and MuViT, we train in a single stage from scratch, without pretraining.

On BetaSeg~\cite{betaseg}, \epssegformer learns from about 10{,}000 labeled pixels in three FIB-SEM volumes of pancreatic $\beta$ cells and segments a slice of a fourth, unseen cell with an average Dice of 0.91: 0.13 above \epsseg trained on the same pixels and only 0.02 below a U-Net trained on all 670 million in-cell voxels, about 67{,}000 times our labels (\cref{tab:comparison}). It is 0.03 above a fully supervised Vanilla ViT and 0.01 below a fully supervised Segmenter.
On mitochondria, it is the most accurate of all compared methods, including these three fully supervised ones.
The test cell is never seen during training, normalization or checkpoint selection, and although configuration and post-processing were chosen on the evaluation slice, a second slice of the same cell gives the same scores (\cref{sec:res-compare}).
These scores are three-seed means; \cref{sec:limitations} discusses limitations such as inference speed.

%% file: sec/2_related.tex
\section{Related Work}
\label{sec:related}

\paragraph{Organelle segmentation in vEM.}
Whole-cell vEM datasets~\cite{xu2021atlas,betaseg} have driven supervised organelle segmentation, from U-Nets~\cite{unet2015,cicek2016unet3d} trained on extensive manual annotation~\cite{heinrich2021cosem} to generalist models such as MitoNet~\cite{conrad2023mitonet}, which relied on large, partly crowdsourced label sets to generalize across samples.
Pretraining on large unlabeled EM collections~\cite{conrad2021cem500k} and segmentation foundation models adapted to microscopy~\cite{kirillov2023sam,archit2025microsam} reduce the need for labels, but they target transfer and promptable object segmentation; assigning organelle classes to every pixel of a new volume still requires examples of those classes.
We study this setting, in which a few labeled pixels define the classes of interest.

\paragraph{Learning from sparse labels.}
Point~\cite{bearman2016point} and scribble~\cite{lin2016scribblesup} supervision lower the annotation cost for natural images, typically by combining the sparse labels with priors on objectness or appearance, and 3D U-Net~\cite{cicek2016unet3d} learns dense volumetric segmentation from a few annotated slices.
In bioimage analysis, ilastik~\cite{berg2019ilastik} and Labkit~\cite{arzt2022labkit} train random forests on a fixed bank of filter features from sparse brush strokes, which is fast and interactive.
MAESTER~\cite{maester2023} learns masked-autoencoder representations of EM without labels and segments by clustering them.
\epsseg~\cite{epsseg2025} trains a hierarchical variational autoencoder with center-region masking and an inpainting loss, shapes its latent space with sparse-label contrastive learning and a Gaussian mixture prior, and predicts classes with an MLP head.
\epssegformer keeps the sparse-label setting and the center inpainting of \epsseg, but replaces its encoder with a multi-resolution transformer.

\paragraph{Patch classifiers with co-centered views.}
Classifying a pixel from a window centered on it is an old recipe in EM: Ciresan et al.~\cite{ciresan2012membranes} segment neuronal membranes with a CNN on the raw pixels around each pixel, trained on dense labels.
Later work adds context with views of several sizes around one center: Farabet et al.~\cite{farabet2013scene} use a multi-scale network on an image pyramid, and DeepMedic~\cite{kamnitsas2017deepmedic} two pathways on a normal and a downsampled patch.
\epssegformer follows this lineage with a multi-resolution transformer and sparse labels, and it hides the center pixel that these classifiers read.

\paragraph{Context and scale in vision transformers.}
Vision Transformers~\cite{dosovitskiy2021vit} relate all tokens through attention, but their cost grows quadratically with the number of tokens, so a large field of view at full resolution is expensive.
Segmenter~\cite{strudel2021segmenter} labels the patch tokens of a plain ViT with a mask transformer. Hierarchical designs build multi-scale feature pyramids from an input at a single resolution~\cite{liu2021swin,fan2021mvit,xie2021segformer}, CrossViT~\cite{chen2021crossvit} fuses tokens from patches of two sizes, and HIPT~\cite{chen2022hipt} aggregates nested regions of gigapixel pathology slides level by level.
MuViT~\cite{muvit2026} instead encodes crops of one image at several physical resolutions with a single encoder, aligns them with rotary position embeddings~\cite{rope,heo2024ropevit} in shared world coordinates, and pretrains with a multi-resolution MAE.
We adopt its encoder, center every view on the labeled pixel, and train directly from sparse labels.

\paragraph{Blind spots and shortcuts.}
Hiding the part of the input that a prediction is about is a recurring self-supervised principle, from context encoders~\cite{pathak2016context} and masked autoencoders~\cite{mae} to blind-spot denoisers that predict a pixel from its neighbors~\cite{krull2019noise2void,batson2019noise2self}.
As shortcut learning shows~\cite{geirhos2020shortcut}, a network tends to fit its objective with the easiest available cue, which makes hiding the input an important design choice.
We carry the blind spot over to classification and hide the token of the labeled pixel in every view, so that its label must be predicted from context.

%% file: sec/3_method.tex
\section{Method}\label{sec:method}

Given EM slices with sparse pixel labels, we predict the class of every pixel inside a cell unseen during training (the in-cell pixels); the cell mask is given (\cref{sec:data}).
\epssegformer{} (\cref{fig:muvit}) classifies a center pixel from $L=3$ views centered on it, whose tokens form one sequence for a transformer encoder; in training the center pixel is a labeled pixel, at inference every in-cell pixel in turn.
Its encoder and decoders build on MuViT~\cite{muvit2026} (differences in Supp.~Tab.~\suppref{tab:muvit_diff}); MuViT denotes our variant with one transformer decoder per view, and \muvitenc{} the variant with a single CNN decoder on the finest view (\cref{sec:training}).
Unlike MuViT, which pretrains with masked reconstruction, we train from scratch in one stage, center every view on the target pixel, and always drop its token.

\subsection{Multi-Resolution Tokens}\label{sec:views}

For a center pixel $p$, we sample $L$ views of the slice and split each into tokens.

\paragraph{Views.}
Let $I$ be the normalized slice (Supp.~Section~\suppref{sec:appendix-data}), extended beyond its border with zeros, the mean intensity (border padding). Without augmentation, view $\ell\in\{1,\dots,L\}$ is the $W\times W$ grid of samples of \cref{eq:views}, centered on $p$ with integer stride $s_\ell$ ($s_1=1$, increasing):
\begin{equation}\label{eq:views}
X^{(\ell)}[i]=I\!\left(p+s_\ell\, i\right),\quad i\in\{-h,\dots,h\}^2,\; W=2h+1.
\end{equation}
All views have $W^2$ samples, but view $\ell$ spans about $s_\ell W$ slice pixels, so a larger $\ell$ gives a wider field of view at a coarser resolution. Coarse views are unsmoothed strided subsamples, and every view's center sample ($i=0$) is $I(p)$.

\paragraph{Tokens.}
Each view is split into $N=(2n+1)^2$ non-overlapping $P\times P$ patches, so $W=P(2n+1)$; since $W$ is odd, so is $P$. Patches are indexed by $j\in\{-n,\dots,n\}^2$, and the patch $j=0$ is centered on $p$ in every view. Each patch is flattened, projected to the model width $d$ by one learned linear layer shared by all views, and added to a learned view embedding; the sum is the token $t^{(\ell)}_j$. Each token carries the slice coordinates $u^{(\ell)}_j=p+s_\ell P\,j$ of its patch center, so all views share one coordinate system, and we encode $u$ with two-dimensional axial RoPE~\cite{rope} (scaling and frequencies in Supp.~Tab.~\suppref{tab:muvit_diff}).

\subsection{Classifying a Hidden Center}\label{sec:hidden}

\paragraph{Rationale.}
A label supervises a single pixel, and that pixel lies in the center token. A classifier that reads this token can fit the labels from the local appearance around $p$ alone, and nothing in the classification loss prevents it. In training, we hide the center token by dropping it from the encoder input, in the spirit of blind-spot denoising~\cite{krull2019noise2void}: the tokens around $p$ cannot receive its content through attention, so the head must rely on them. The neighboring tokens stay visible unless dropped at random, and blur augmentation (\cref{sec:training}) can carry some of the intensity of $p$ into them. \Cref{sec:res-scales} tests this choice.

\paragraph{Dropping Tokens.}
Every view is centered on $p$ (\cref{sec:views}), so dropping only the finest center token would leave $p$ visible in the coarser views, except by chance. We therefore drop the center token of every view and, in each view, $\lfloor\rho N\rfloor$ further tokens drawn uniformly at random, independently for every sample and view; $\rho$ is the drop ratio (\cref{sec:impl}). The dropped set $\mathcal M$ of a training sample contains the center tokens $(\ell,0)$ for $\ell=1,\dots,L$ and the random tokens; the visible set $\mathcal V$ contains all others.

\paragraph{Encoder.}
The encoder is a pre-norm transformer without a class token, since the head reads patch tokens. It maps the visible tokens $t_{\mathcal V}$ of all views, with their coordinates $u_{\mathcal V}$, to $z_{\mathcal V}=f_\theta(t_{\mathcal V},u_{\mathcal V})$, and every visible token attends to all visible tokens of all views. As in MAE~\cite{mae}, dropped tokens never enter the encoder; after encoding, a learned mask token $m$ replaces the output at every dropped position, $z_{\mathcal M}=m$. The head and decoders read all outputs $z^{(\ell)}_j$.

\paragraph{Context-Only Head.}
The segmentation head (S.H.) predicts the class of $p$ from the encoder outputs around the center token of the finest view. Let $\mathcal N_k=\{j\in\{-n,\dots,n\}^2:\lVert j\rVert_\infty\le (k-1)/2\}\setminus\{0\}$ be the $k\times k$ neighborhood of the center token, without the center token itself, where $k$ is odd and $3\le k\le 2n+1$. The prediction (\cref{eq:head}) is
\begin{equation}\label{eq:head}
\hat y=\mathrm{softmax}\Bigl(g_\phi\Bigl(\mathop{\mathrm{concat}}_{j\in\mathcal N_k}\mathrm{LN}\bigl(z^{(1)}_j\bigr)\Bigr)\Bigr).
\end{equation}
Here $\mathrm{LN}$ is a shared LayerNorm and $g_\phi$ a two-layer MLP with a GELU that outputs one score per class.

\subsection{Decoders and One-Stage Training}\label{sec:training}

\paragraph{Decoders.}
Decoder $\ell$ reconstructs the dropped patches of view $\ell$ with two transformer layers and RoPE: the first cross-attends from the tokens of view $\ell$ to all $L\cdot N$ tokens, with the mask token $m$ at the dropped positions, and the second self-attends within view $\ell$. A linear layer maps each token to its $P^2$ pixels and gives the reconstruction $\hat X^{(\ell)}$ (layers in Supp.~Section~\suppref{sec:appendix-muvit}). At inference, the decoders take no part in the prediction (\cref{sec:inference}). The single CNN decoder of \muvitenc{} mixes patches only through one $3{\times}3$ convolution, so it cannot infer a dropped patch from its wider context (Supp.~Section~\suppref{sec:appendix-muvit}).

\paragraph{Objective.}
For a labeled pixel $p$ with label $y$, the loss (\cref{eq:loss}) is the unweighted sum of a mean squared reconstruction error and the cross-entropy of the head's prediction:
\begin{equation}\label{eq:loss}
\mathcal L=\frac1L\sum_{\ell=1}^{L}\frac1{|\Pi_\ell|}\sum_{i\in\Pi_\ell}\bigl(\hat X^{(\ell)}[i]-X^{(\ell)}[i]\bigr)^2\;-\;\log\hat y_y .
\end{equation}
Here $\Pi_\ell$ is the set of samples of view $\ell$ under its dropped tokens, and the target $X^{(\ell)}$ is the augmented view $\ell$. For \muvitenc{}, only the term $\ell=1$ remains.

\paragraph{Augmentation.}
Training views are augmented geometrically (flips, quarter turns, a small rotation, an elastic field) and photometrically (gamma, contrast, brightness, blur, noise; parameters in Supp.~Section~\suppref{sec:appendix-impl}). One random draw is shared by all views of a sample, so they stay aligned. The geometric transformation fixes $p$, so every view stays centered on $p$ and the label stays valid.

\subsection{Inference and Post-Processing}\label{sec:inference}

\paragraph{Dense Inference.}
The cell mask selects the pixels to classify and is not a model input. Each center pixel $p$ gets its own $L$ views, built as in training (\cref{eq:views}) with border padding and without augmentation. Nothing is dropped ($\mathcal M=\emptyset$), so the encoder uses all available information: the center token, which training always drops, is visible and can pass information to the neighborhood tokens through attention. The head reads the same $\mathcal N_k$ as in training. The prediction is $\arg\max_c\hat y_c(p)$ (\cref{eq:head}).

\paragraph{Test-Time Augmentation (TTA3).}
TTA3 classifies the slice and its horizontal and vertical flips, each as above, maps the predictions back, and averages the three class-probability vectors before the argmax; no parameter is tuned on the evaluation slice.

\paragraph{Post-Processing.}
A final step refines the class map from the class probabilities (averaged under TTA3) and the cell mask, changing only in-cell pixels. It decides the classes from the largest structure to the smallest (Nucleus, Mitochondria, Granule) on Gaussian-smoothed probabilities, gives the rest to Uncategorized, and splits fused granules with a distance-transform watershed. The rules and parameters are in Supp.~Section~\suppref{sec:appendix-postproc}; they were tuned as in \cref{sec:postproc}.

%% file: sec/4_experiments.tex
\section{Data}
\label{sec:data}

\paragraph{Dataset and Classes.}
We use the BetaSeg focused ion beam scanning electron microscopy (FIB-SEM) volumes of primary mouse pancreatic $\beta$ cells~\cite{betaseg}: four high-glucose cells, c1--c3 and high\_c4, with isotropic 16\,nm voxels. We process them as 2D XY slices of 16\,nm pixels (Supp.~Tab.~\suppref{tab:data}). Each pixel inside the segmented cell (in-cell pixel) belongs to one of four classes merged from the BetaSeg organelle masks: Nucleus, Granule (insulin secretory granules), Mitochondria, and Uncategorized, a labeled class that holds the rest of the cell, including Golgi, microtubules and plasma membrane. Pixels outside the cell are context only: they carry no label and are neither classified nor scored. The cell mask comes from the ground truth, also at test time (\cref{sec:inference}).

\paragraph{Sparse Labels.}
Following $\varepsilon$-Seg~\cite{epsseg2025}, we simulate sparse labels by sampling single pixels from the dense ground truth. From each training slice, we draw pixels uniformly at random within each class: one per class, and two for Nucleus in the slices that contain a nucleus (about 47\%). Each class therefore holds 24--26\% of the labels, against 7--58\% of the in-cell voxels (Supp.~Tab.~\suppref{tab:data}). We use 10,029 labels, 3.8 per slice, or $1.5\times10^{-5}$ of the in-cell voxels of c1--c3.

\paragraph{Training and Validation.}
The c1--c3 slices that contain the cell are shuffled with a fixed seed and split 85/15 into 2,611 training and 463 validation slices. The validation slices carry their own simulated labels (Supp.~Tab.~\suppref{tab:data}). They are random slices of the training cells, 16\,nm from their neighbors, so validation loss selects the checkpoint but does not estimate generalization.

\paragraph{Evaluation Slice.}
Cell~high\_c4 is unseen by training: it contributes no labels, normalization statistics or checkpoint selection.
Following \epsseg, we evaluate on slice~627 of high\_c4 ($545{\times}1082$ pixels; slices are counted from 1), to compare with the published baselines without retraining them.
We score all of its 317,438 in-cell pixels with the Avg.\ Dice (\cref{sec:metrics}).
As there is no separate test set, we made development choices on this slice, including the final configuration (\cref{sec:configs}), and use slice~577 of the same cell (50 slices away) as a check (\cref{sec:res-compare}); \cref{sec:res-select,sec:res-final} bound the resulting optimism.

%% file: sec/5_experiments.tex
\section{Experiments}\label{sec:experiments}

\paragraph{Final Model.}\label{sec:configs}
The final model is \epssegformer{} with the MuViT architecture. It has $L=3$ views of $85{\times}85$ samples ($W=85$), each cut into $5{\times}5$ patches ($P=5$), which gives $N=17\times17=289$ patches per view and $L\cdot N=867$ tokens. The scales are $\{1,4,8\}$ (the strides $s_\ell$ of \cref{sec:views}), so the views span 85, 340 and 680 slice pixels. The S.H.\ kernel is $5{\times}5$ ($k=5$). The encoder has 14 transformer layers of width $d=192$, one attention head and an MLP ratio of 2; the MLP $g_\phi$ of \cref{eq:head} has a hidden width of 256. The model has 7.25M parameters; 5.35M are used at inference, and the three decoders (1.91M) are not. \muvitenc{} has the same 5.35M at inference.

\paragraph{Training.}\label{sec:impl}
We train with AdamW at a constant learning rate of $10^{-4}$ and batch size 96; an epoch is 100 iterations, about one pass over the 10,029 labels of \cref{sec:data}. The drop ratio $\rho$ falls from 0.70 to 0.25 within 15 epochs. The ablations train for 200 epochs and the three-seed runs stop early on the validation loss. Schedule and augmentation parameters are in the supplement (Supp.~Section~\suppref{sec:appendix-impl}).

\paragraph{Checkpoint.}\label{sec:ckpt}
We keep the checkpoint with the lowest validation loss, that is, the loss of \cref{eq:loss} on the validation labels of c1--c3 at the current drop ratio. The reconstruction term differs between the architectures (the mean over the views for MuViT, the finest view for \muvitenc{}), so their validation losses are not comparable.

\paragraph{Ablations.}
Unless a table says otherwise, the ablations (\cref{sec:res-augpad,sec:res-scales}) use the scales $\{1,2,4\}$ and drop only the center token of the finest view.

\paragraph{Configuration Selection.}
Both architectures are trained with three seeds at views of $65{\times}65$ and $85{\times}85$ (507 and 867 tokens) and S.H.\ kernels of $3{\times}3$ and $5{\times}5$, which gives eight configurations (\cref{sec:res-select}). They use the recipe of the final model: scales $\{1,4,8\}$, augmentation, border padding, the center token of every view dropped, and early stopping.

\paragraph{Baselines.}\label{sec:baselines}

\Cref{tab:comparison} compares \epssegformer{} with eight methods, in ten rows, that differ in supervision. Vanilla HVAE, Han et al. and MAESTER are self-supervised. Labkit, \epsseg{} (10k and 100k labels) and a U-Net trained on three fully labeled slices (946{,}002 labeled pixels) are sparsely supervised. A second U-Net, a Vanilla ViT and Segmenter are fully supervised, on all 669{,}571{,}304 labeled pixels of the three training cells. The numbers of the first six methods are taken from the \epsseg{} paper; all methods are scored on slice~627 of cell high\_c4, and \epsseg{} (10k) and \epssegformer{} are trained on the same pixel coordinates.

\input{fig/qualitative}

\paragraph{Metric.}\label{sec:metrics}
We report the average Dice (Avg.\ Dice), the mean of the per-class Dice over the four classes, on all in-cell pixels of the evaluation slice (ablations without border padding score slightly fewer, \cref{sec:res-augpad}).
Each model gives four scores, the columns Raw (one forward pass), TTA3, Post and TTA3 + Post of \cref{tab:config_selection}; TTA3 and post-processing are part of \epssegformer{}, so TTA3 + Post is our headline score, and other tables are raw unless marked.

\paragraph{Post-Processing.}\label{sec:postproc}
Post-processing was tuned on the evaluation slice: its Gaussian widths come from ground-truth object sizes (Supp.~Section~\suppref{sec:appendix-postproc}), and the remaining choices were ranked on the left half of the slice with earlier models, without TTA3, before the final model was evaluated.
We therefore report its gain on the right half, rank configurations by their raw score, and check it on a second slice (\cref{sec:res-final}); TTA3 has no tuned parameter.

\paragraph{Seeds and Uncertainty.}\label{sec:seeds}
Ablations are single runs (seed~42); the eight configurations of \cref{tab:config_selection} use three seeds (54, 74, 313), which change initialization, sample order, dropped tokens and augmentation but not the label set.
We report mean $\pm$ SD (per seed and class in Supp.~Tabs.~\suppref{tab:config_full} and~\suppref{tab:perclass}) and do not read differences below 0.005 between three-seed means, or 0.009 between single runs, as real: each is one standard error of the difference, a noise screen, not a significance test (Supp.~Section~\suppref{sec:appendix-threshold}).

%% file: fig/qualitative.tex
\renewcommand{\qgcolumns}{Input/raw/, Ground truth/gt/, \epssegformer/eps-Segformer/10k, MAESTER/maester/{No labels}, U-Net/unet/670M, \epsseg/eps-Seg/10k}
\renewcommand{\qgroot}{figs/qual}
\renewcommand{\qgperrow}{3}      %
\renewcommand{\qgzar}{0.6}      %
\renewcommand{\qgframelw}{0.4pt}
\renewcommand{\qgzoomsep}{0.2pt}
\renewcommand{\qgrowsep}{0.6pt}
\renewcommand{\qgoverlay}{1}   %
\renewcommand{\qgAR}{0.5037}   %
\providecommand{\qgchip}[1]{{\setlength{\fboxsep}{0pt}\setlength{\fboxrule}{0.3pt}\fcolorbox{black!70}{#1}{\rule{0pt}{0.62em}\rule{0.62em}{0pt}}}}
\providecommand{\tbd}{\textcolor{red}{\textit{0.0000}}}
\providecommand{\tbdA}{\tbd\hphantom{\,{\scriptsize$\pm$0.0000}}}%
\begin{figure*}[t]
  \centering
  \qualrow[0]{sample01}{0.5545,0.3894,0.1183}{0.1386,0.5526,0.1183}
  \caption{\textbf{Qualitative Results} on slice 627 (cell high\_c4). The panels compare the input, ground truth, and predictions of \epssegformer{} and of three methods from \cref{tab:comparison} (MAESTER, U-Net and \epsseg{}) displaying the full slice above two magnified regions. Labels indicate the method and labeled training pixel count. The \epssegformer{} result (random seed 74) uses TTA3 (test-time augmentation over the slice and its two flips) and post-processing, and achieves an Avg. Dice of 0.9124. Raw (0.8938) is the prediction before both. Classes: \qgchip{qgNucleus}~nucleus, \qgchip{qgGranules}~granule, \qgchip{qgMito}~mitochondria, \qgchip{qgOther}~uncategorized.}
  \label{fig:qual_grid}
  \par\vspace{6pt}
  \begingroup%
  \captionof{table}{\textbf{Quantitative Analysis.} Dice per class and average on slice 627 (cell high\_c4). Baseline results (Vanilla HVAE~\cite{prakash2022hdn}, Han et al.~\cite{han2022selfsup}, MAESTER~\cite{maester2023}, Labkit~\cite{arzt2022labkit}, U-Net~\cite{unet2015}, Vanilla ViT~\cite{dosovitskiy2021vit}, Segmenter~\cite{strudel2021segmenter}, and \epsseg{}) are cited from the \epsseg{} paper~\cite{epsseg2025}. \epsseg{} (10k) and \epssegformer{} trained on identical pixel coordinates. The two \epssegformer{} rows report mean $\pm$ SD over three seeds for raw predictions and those with TTA3 + post-processing (per seed: Supp.~Tabs.~\suppref{tab:config_full} and~\suppref{tab:perclass}). \Cref{fig:qual_grid} shows \epssegformer{} (10k), MAESTER, \epsseg{} (10k) and fully supervised U-Net (670M).}
  \label{tab:comparison}
  \tabstylewide
  \setlength{\tabcolsep}{3.5pt}
  \providecommand{\nl}[1]{\hfill[#1]}%
  \begin{tabular}{l >{}p{4.1cm} ccccc}
  \toprule
  Supervision & Method\hfill[\# Labels] & \qgchip{qgNucleus}\,Nucleus & \qgchip{qgGranules}\,Granule & \qgchip{qgMito}\,Mitochondria & \qgchip{qgOther}\,Uncategorized & Avg.\ Dice \\
  \midrule
         & Vanilla HVAE\textsuperscript{*} \nl{no labels} & 0.55 & 0.34 & 0.13 & 0.44 & 0.37 \\
         & Han et al.\textsuperscript{*} \nl{no labels} & --   & --   & --   & --   & 0.66 \\
  \multirow{-3}{*}{Self-Supervised} & MAESTER\textsuperscript{*} \nl{no labels} & 0.95 & 0.56 & 0.79 & 0.84 & 0.79 \\
  \midrule[0.3pt]
         & Labkit \nl{10k} & 0.44 & 0.68 & 0.61 & 0.85 & 0.65 \\
         & \epsseg{} \nl{10k}          & 0.95 & 0.69 & 0.69 & 0.79 & 0.78 \\
         & \epsseg{} \nl{100k}         & 0.96 & 0.82 & 0.86 & 0.91 & 0.89 \\
         & U-Net (on 3 slices) \nl{946k} & 0.96 & 0.78 & 0.66 & 0.90 & 0.83 \\
         & \epssegformer{} (ours, raw) \nl{10k} & 0.9739\,{\scriptsize$\pm$0.0010} & 0.7801\,{\scriptsize$\pm$0.0145} & 0.9023\,{\scriptsize$\pm$0.0056} & 0.9048\,{\scriptsize$\pm$0.0085} & 0.8903\,{\scriptsize$\pm$0.0073} \\
  \multirow{-6}{*}{\shortstack[l]{Sparsely\\Supervised}} & \cellcolor{black!9}\textbf{\epssegformer{} (ours)} \nl{10k} & \cellcolor{black!9}0.9778\,{\scriptsize$\pm$0.0019} & \cellcolor{black!9}0.8120\,{\scriptsize$\pm$0.0084} & \cellcolor{black!9}0.9216\,{\scriptsize$\pm$0.0052} & \cellcolor{black!9}0.9245\,{\scriptsize$\pm$0.0051} & \cellcolor{black!9}0.9090\,{\scriptsize$\pm$0.0048} \\
  \midrule[0.3pt]
         & Vanilla ViT \nl{670M} & 0.98 & 0.77 & 0.87 & 0.91 & 0.88 \\
         & Segmenter \nl{670M} & 0.99 & 0.86 & 0.90 & 0.91 & 0.92 \\
  \multirow{-3}{*}{\shortstack[l]{Fully\\Supervised}} & U-Net \nl{670M} & 0.99 & 0.90 & 0.87 & 0.94 & 0.93 \\
  \bottomrule
  \multicolumn{7}{@{}l}{\footnotesize\rule{0pt}{1.2em}\textsuperscript{*}\,Methods marked with an asterisk use K-Means clustering on latent features to conduct semantic segmentation.}
  \end{tabular}
  \endgroup
\end{figure*}

%% file: sec/5_results.tex
\section{Results}
\label{sec:results}

All scores are Avg.\ Dice on the in-cell pixels of slice~627 of cell high\_c4 (\cref{sec:metrics}), unless a table states otherwise.

\subsection{Comparison With Other Methods}\label{sec:res-compare}

On slice~627 of high\_c4, \epssegformer{} reaches an Avg.\ Dice of $0.9090\pm0.0048$ with TTA3 and post-processing (mean $\pm$ SD over three seeds), and $0.8977\pm0.0056$ with TTA3 only (\cref{tab:config_selection}). \Cref{tab:comparison} lists this mean, and \cref{fig:qual_grid} shows the best seed (0.9124). Granule is the lowest class in every seed (Supp.~Tab.~\suppref{tab:perclass}). On a second slice of the same cell, slice~577, the final configuration scores 0.8910 raw and 0.9080 post-processed, within 0.0007 of its slice-627 scores (Supp.~Section~\suppref{sec:appendix-slice577}). On 17 further slices of the same cell that contain the nucleus class, one seed scores $0.884\pm0.015$ raw and $0.904\pm0.011$ with TTA3 and post-processing (mean $\pm$ SD over slices; Supp.~Section~\suppref{sec:appendix-volume}).

The three-seed mean is about 0.13 above \epsseg{} (10k labels, the same pixels as ours), 0.12 above MAESTER, 0.08 above the U-Net trained on three full slices, 0.02 above \epsseg{} (100k labels; level with our single-pass 0.89), 0.03 above the fully supervised Vanilla ViT, 0.01 below the fully supervised Segmenter and 0.02 below the U-Net trained on dense labels (\cref{tab:comparison}). Our post-processing was tuned on this slice (\cref{sec:res-final}); the \epsseg{} numbers also include post-processing. Per class, Mitochondria has the highest Dice of all methods (0.92, against 0.90 for Segmenter and 0.87 for the dense U-Net and the Vanilla ViT), while on Granule the dense U-Net (0.90) and Segmenter (0.86) stay ahead of it (0.81), and only the Vanilla ViT is lower (0.77).

Errors concentrate near boundaries. In the TTA3 output, 71.0\% of the misclassified pixels (68.5\% to 73.0\% over the three seeds) lie within 3\,px (Euclidean, 48\,nm) of a ground-truth boundary between two classes or between cell and background, a band that holds 31.6\% of the in-cell pixels.

\subsection{Augmentation and Border Padding}\label{sec:res-augpad}

Augmentation raises Avg.\ Dice by $+0.0769$ for MuViT and $+0.0755$ for \muvitenc{}, by far the largest change in \cref{tab:roadmap}.
Without it, validation loss is lowest at epochs 19--37 and the models then overfit; with it, at epochs 91--198.

\input{fig/tab_roadmap}

\input{fig/tab_config_selection}

Border padding completes views that extend beyond the slice, so labels near the border stay in training.
It adds $+0.0165$ Avg.\ Dice for \muvitenc{}, growing with view size, and $+0.0009$ for MuViT (Supp.~Tab.~\suppref{tab:padding}).
Without padding, up to 1,664 of the 10,029 labels are unusable and scoring skips under 2\% of the in-cell pixels; on identical pixels the gains are slightly larger ($+0.0188$ and $+0.0025$; Supp.~Tab.~\suppref{tab:padding_pixels}).
The final $85{\times}85$ views need padding: their coarsest view is wider than the slice is high.

\subsection{Views, Scales, Kernel and Center Token}\label{sec:res-scales}\label{sec:res-kernel}

\paragraph{Coarser Views.}
Adding the coarser views raises Avg.\ Dice in all 18 seed pairs. We retrain the six configurations with a $5{\times}5$ kernel (both architectures, three view sizes) with only the finest view (scale~1), three seeds each and all else equal, and compare them with the scales $\{1,4,8\}$ (Supp.~Tab.~\suppref{tab:scales1_full}). The gain is $+0.0156$ to $+0.0357$ per configuration ($+0.0249$ on average); at the final configuration, Avg.\ Dice rises from $0.8546\pm0.0134$ to $0.8903\pm0.0073$. The single view has a third of the tokens and an eighth of the field of view, so this comparison confounds context with token count.

\paragraph{Scales and Kernel.}
Replacing the scales $\{1,2,4\}$ by $\{1,4,8\}$ changes Avg.\ Dice by $+0.0028$ on average. Seven of the eight changes are positive, but only one ($+0.0110$, MuViT at $27{\times}27$ with a $3{\times}3$ kernel) exceeds the single-run threshold of 0.009, so we claim no clear advantage.
We use $\{1,4,8\}$, which scores no lower on average and whose coarsest view is twice as wide at the same number of tokens.
The best S.H.\ kernel is $5{\times}5$ in three of the four combinations of view size and architecture; the larger kernels $9{\times}9$ and $13{\times}13$ fall $0.02$ to $0.07$ below it (Supp.~Tab.~\suppref{tab:scales_full}).

\paragraph{Hiding the Center Token.}
The center token is hidden so that the head cannot fit a label to the local appearance of the labeled pixel (\cref{sec:hidden}). In a single-view ViT, always dropping it raises the mean Avg.\ Dice by $+0.0141$ and $+0.0035$ in the two drop patterns compared, though not in every setting (Supp.~Tab.~\suppref{tab:masking-summary}). At the final configuration, we repeat the three seeds with the center token of every view kept visible in training; this also gives the segmentation head one more token during training and inference. Avg.\ Dice falls from $0.8903\pm0.0073$ to $0.8826\pm0.0069$ ($-0.0077$) and is lower in each of the three seeds (Supp.~Tab.~\suppref{tab:center_full}).

\subsection{Configuration Selection With Three Seeds}\label{sec:res-select}

MuViT with $85{\times}85$ views, $5{\times}5$ patches and a $5{\times}5$ kernel ranks first among the eight configurations in all four columns of \cref{tab:config_selection}, but by at most 0.0031, below the 0.005 threshold of \cref{sec:seeds} and below its own seed SD; all eight lie within 0.0126 raw.
We use it because it ranks first everywhere, knowing that it was selected on this slice.
Paired by seed, the $5{\times}5$ kernel leads $3{\times}3$ by 0.0055 raw, while architecture and view size do not separate the runs (Supp.~Section~\suppref{sec:appendix-pairs}).

\subsection{Test-Time Augmentation and Post-Processing}\label{sec:res-final}

TTA3 raises Avg.\ Dice in all eight configurations, by 0.0062 to 0.0078 raw (0.0074 for the final model) and by 0.0011 to 0.0027 after post-processing (0.0017). Post-processing raises it by 0.0167 to 0.0237 on the raw output (0.0170) and by 0.0113 to 0.0187 on the TTA3 output (0.0113).

On the TTA3 output, post-processing raises Avg.\ Dice by $+0.0125\pm0.0009$ on the left half, where the pipeline was ranked, and by $+0.0116\pm0.0016$ on the right half, which played no part in the ranking (mean $\pm$ SD over three seeds). On the right half every class improves in all three seeds ($+0.0019$ to $+0.0276$), so the gain is not an artifact of the ranking. The three Gaussian widths did see the whole slice, however, so the absolute post-processed scores remain tuned on it. On slice~577, which was not used for tuning, post-processing raises the raw Avg.\ Dice of the final configuration by $+0.0170$, as on slice~627 (Supp.~Tab.~\suppref{tab:slice577}).

%% file: fig/tab_roadmap.tex
\begin{table}[t]
\centering
\caption{\textbf{Effect of Augmentation and Border Padding.} Avg.\ Dice of \epssegformer{} with data augmentation and border padding, for the MuViT and \muvitenc{} architectures. Values are averaged across four configurations of view/patch size and segmentation head kernel size (one training run per setting; see Supp.~Tabs.~\suppref{tab:aug_ctx} and \suppref{tab:padding}). Rows are cumulative (each adds the feature to the one above), with $\Delta$ the gain in Avg.\ Dice over the previous row.}
\label{tab:roadmap}
\tabstyle
\setlength{\tabcolsep}{2.5pt}
\begin{tabular}{@{}l cc@{\hspace{6pt}}cc@{}}
\toprule
 & \multicolumn{2}{c}{MuViT} & \multicolumn{2}{c}{\muvitenc} \\
\cmidrule(lr){2-3}\cmidrule(l){4-5}
Step & Avg.\ Dice & $\Delta$ & Avg.\ Dice & $\Delta$ \\
\midrule
1) No Aug., no padding & 0.8009 & -- & 0.7904 & -- \\
2) + Aug. & 0.8778 & $+0.0769$ & 0.8659 & $+0.0755$ \\
3) + Aug. + Border padding & 0.8787 & $+0.0009$ & 0.8824 & $+0.0165$ \\
\bottomrule
\end{tabular}
\end{table}

%% file: fig/tab_config_selection.tex
\begin{table}[t]
\centering
\caption{\textbf{Eight Configurations Compared Over Three Training Seeds: Architecture, View Size and Segmentation Head (S.H.)\ Kernel.} Avg.\ Dice is reported as mean (with standard deviation) across three seeds (same data, varying initialization and sample order). Raw is a single forward pass without TTA3 or post-processing. +TTA3 adds test-time augmentation (\cref{sec:inference}), +Post adds slice-tuned post-processing (\cref{sec:metrics}), and +TTA3 +Post applies both. Bold indicates the best of the eight configurations, underline the best \muvitenc{}, and shading the final configuration. For the per-seed values, see Supp.~Tab.~\suppref{tab:config_full}.}
\label{tab:config_selection}
\tabstyle
\setlength{\tabcolsep}{3.0pt}\renewcommand{\arraystretch}{1.0}
\begin{tabular}{ll cccc}
\toprule
\makecell[l]{View\,/\,patch\\[-1pt](px)} & \makecell{S.H.\\[-1pt]kernel} & Raw & +TTA3 & +Post & \makecell{+TTA3\\[-1pt]+Post} \\
\midrule
\multicolumn{6}{l}{\textbf{MuViT}} \\
$65$ / $5$ & \multirow{2}{*}{$3{\times}3$} & 0.8819 & 0.8889 & 0.9056 & 0.9076 \\[-1.5pt]
{\scriptsize(507 tokens)} &  & {\scriptsize($\pm$0.0081)} & {\scriptsize($\pm$0.0094)} & {\scriptsize($\pm$0.0060)} & {\scriptsize($\pm$0.0069)} \\[0.5pt]
 & \multirow{2}{*}{$5{\times}5$} & 0.8837 & 0.8913 & 0.9041 & 0.9068 \\[-1.5pt]
 &  & {\scriptsize($\pm$0.0048)} & {\scriptsize($\pm$0.0048)} & {\scriptsize($\pm$0.0073)} & {\scriptsize($\pm$0.0070)} \\[1.5pt]
$85$ / $5$ & \multirow{2}{*}{$3{\times}3$} & 0.8808 & 0.8871 & 0.9038 & 0.9049 \\[-1.5pt]
{\scriptsize(867 tokens)} &  & {\scriptsize($\pm$0.0041)} & {\scriptsize($\pm$0.0051)} & {\scriptsize($\pm$0.0065)} & {\scriptsize($\pm$0.0075)} \\[0.5pt]
\rule{0pt}{7.6pt}& \cellcolor{black!9} & \cellcolor{black!9}\textbf{0.8903} & \cellcolor{black!9}\textbf{0.8977} & \cellcolor{black!9}\textbf{0.9073} & \cellcolor{black!9}\textbf{0.9090} \\[-1.5pt]
 & \cellcolor{black!9}\multirow{-2}{*}{$5{\times}5$} & \cellcolor{black!9}{\scriptsize($\pm$0.0073)} & \cellcolor{black!9}{\scriptsize($\pm$0.0056)} & \cellcolor{black!9}{\scriptsize($\pm$0.0050)} & \cellcolor{black!9}{\scriptsize($\pm$0.0048)} \\[0.5pt]
\midrule
\multicolumn{6}{l}{\textbf{\muvitenc}} \\
$65$ / $5$ & \multirow{2}{*}{$3{\times}3$} & 0.8858 & 0.8921 & 0.9036 & 0.9048 \\[-1.5pt]
{\scriptsize(507 tokens)} &  & {\scriptsize($\pm$0.0080)} & {\scriptsize($\pm$0.0073)} & {\scriptsize($\pm$0.0067)} & {\scriptsize($\pm$0.0055)} \\[0.5pt]
 & \multirow{2}{*}{$5{\times}5$} & \underline{0.8884} & \underline{0.8946} & \underline{0.9051} & \underline{0.9063} \\[-1.5pt]
 &  & {\scriptsize($\pm$0.0024)} & {\scriptsize($\pm$0.0039)} & {\scriptsize($\pm$0.0036)} & {\scriptsize($\pm$0.0046)} \\[1.5pt]
$85$ / $5$ & \multirow{2}{*}{$3{\times}3$} & 0.8777 & 0.8842 & 0.8967 & 0.8985 \\[-1.5pt]
{\scriptsize(867 tokens)} &  & {\scriptsize($\pm$0.0105)} & {\scriptsize($\pm$0.0109)} & {\scriptsize($\pm$0.0096)} & {\scriptsize($\pm$0.0103)} \\[0.5pt]
 & \multirow{2}{*}{$5{\times}5$} & 0.8858 & 0.8936 & 0.9034 & 0.9055 \\[-1.5pt]
 &  & {\scriptsize($\pm$0.0010)} & {\scriptsize($\pm$0.0010)} & {\scriptsize($\pm$0.0020)} & {\scriptsize($\pm$0.0028)} \\[1.5pt]
\bottomrule
\end{tabular}
\end{table}

%% file: sec/6_conclusion.tex
\section{Discussion and Conclusion}
\label{sec:conclusion}

With about 10{,}000 labeled pixels, \epssegformer, a ViT with 5.35M parameters at inference, trained from scratch in one stage, reaches an Avg.\ Dice score that lies 0.13 above \epsseg trained on the same pixels and only 0.02 below a U-Net trained on all 670 million in-cell voxels, about 67{,}000 times as many labels (\cref{tab:comparison}).
The coarser views account for much of this: keeping only the finest view lowers Avg.\ Dice in every configuration, by 0.016 to 0.036, whereas the other design choices change it by less than the variation between seeds (\cref{sec:res-scales,sec:res-select}).

Hiding the center token is the part of the method that most directly addresses learning from single pixels (\cref{sec:hidden}).
In early single-view runs, always dropping it scored higher in both comparisons (Supp.~Tab.~\suppref{tab:masking-summary}), and at the final configuration, keeping it visible in every view lowers Avg.\ Dice in all three seeds, by 0.0077 on average (\cref{sec:res-scales}).

The per-class scores show where balanced sparse labels help.
\epssegformer is more accurate than the dense U-Net, the Vanilla ViT and Segmenter on Mitochondria (0.92 against 0.87, 0.87 and 0.90), the class that the dense ground truth represents least: Mitochondria cover 7\% of the in-cell voxels of the training cells but receive about a quarter of our sparse labels, which are balanced across classes (Supp.~Tab.~\suppref{tab:data}).
We attribute the gain to this balance combined with sufficient context: \epsseg, trained on the same balanced labels, reaches only 0.69 on Mitochondria, so the labels alone do not explain it.
Consistent with this, Granule, whose share hardly changes (22\% of the voxels, a quarter of the labels), shows no such gain, and the dense U-Net and Segmenter stay ahead on it (0.90 and 0.86 against 0.81).

\paragraph{Limitations.}\label{sec:limitations}
Like the baselines, we evaluate on slice~627 of one held-out cell from a single dataset, on which we also selected the configuration and post-processing; a second slice of the same cell gives the same scores within 0.0040 for every configuration (Supp.~Section~\suppref{sec:appendix-slice577}), but other cells remain untested.
\Cref{tab:comparison} compares three-seed means with single values for the baselines, and most ablations are single runs.
The simulated labels are exact and class-balanced, whereas human annotators err and may not balance classes.
Finally, inference runs one forward pass per pixel; this parallelizes easily but makes whole volumes expensive: about 1.6 GPU-hours per slice on a V100, and 41 GPU-days for the in-cell voxels of one cell (Supp.~Section~\suppref{sec:appendix-impl}).

\paragraph{Outlook.}
We find three future directions particularly interesting.
First, dense inference must become faster before whole volumes can be segmented routinely, for instance by dropping the decoders and classifying many center pixels from shared views.
Second, the isotropic voxels of FIB-SEM invite context along $z$, for instance through views in orthogonal planes or volumetric tokens.
Third, sparse labels lend themselves to active learning, in which the model asks the annotator for new labels where its current predictions are least certain.
This requires uncertainty estimates that can be trusted, which motivates a variational multi-resolution encoder, in the spirit of the hierarchical VAE of \epsseg.

\paragraph{Conclusion.}
A multi-resolution transformer, trained from scratch on single labeled pixels with the center hidden from every view, closes most of the gap between sparse and dense supervision for organelle segmentation in FIB-SEM on an unseen cell.
Broader evaluation and faster inference are the next steps toward segmenting whole cells from about ten thousand clicks.

%% file: sec/6_llm_statement.tex
\section*{LLM Usage Statement}
\label{sec:llm}
We used an LLM-based assistant (Claude, Anthropic) for coding, typesetting and writing.
\emph{Code and experiments:} writing and debugging training, evaluation and plotting scripts, and cross-checking reported numbers against experiment outputs.
\emph{Figures and tables:} plotting code and \LaTeX{} layout of tables, figures and floats; the electron microscopy images and segmentations shown are unedited data and outputs of our models, and no figure content is AI-generated.
\emph{Main text and supplement:} drafting and editing of text, condensing the text to the page limit, grammar, literature search, and verification of bibliography entries.
\emph{Ideation:} no research ideas; only suggestions for how to frame some aspects of the paper clearly within limited space.
The authors designed all experiments, verified all content, and take full responsibility for it.

%% file: sec/7_appendix.tex
This supplement has five parts, the first four in the order in which the main paper uses them.
\begin{list}{}{\setlength{\leftmargin}{1em}\setlength{\itemindent}{-1em}\setlength{\labelwidth}{0pt}\setlength{\labelsep}{0pt}%
  \setlength{\topsep}{2pt}\setlength{\partopsep}{0pt}\setlength{\itemsep}{1pt}\setlength{\parsep}{0pt}\setlength{\listparindent}{0pt}}
\item \textbf{Part~\ref{sec:appendix-setup}} gives the architecture, the data and the training details (Section~\mainref{sec:method}, \mainref{sec:data} and \mainref{sec:impl}).
\item \textbf{Part~\ref{sec:appendix-final}} gives the results of the final model: the per-class Dice of all seeds (Table~\mainref{tab:comparison} and Section~\mainref{sec:res-compare}) and its predictions on the whole slice of Figure~\mainref{fig:qual_grid}, before and after post-processing (Section~\mainref{sec:inference} and \mainref{sec:res-final}).
\item \textbf{Part~\ref{sec:appendix-ablations}} gives the single-run ablations of the design choices: dropping the center token, augmentation, border padding, coarser views, crop scales, the S.H.\ kernel and the view size (Section~\mainref{sec:hidden}, \mainref{sec:res-augpad} and \mainref{sec:res-scales}).
\item \textbf{Part~\ref{sec:appendix-selection}} gives the three seeds behind the choice of the final configuration (Section~\mainref{sec:res-select}).
\item \textbf{Part~\ref{sec:appendix-limitations}} gives the limitations in detail, which the Discussion and Conclusion of the main paper summarizes (Section~\mainref{sec:limitations}).
\end{list}
\noindent Part~\ref{sec:appendix-final} alone covers the headline result; Parts~\ref{sec:appendix-ablations} and~\ref{sec:appendix-selection} hold the evidence for the design choices and for the configuration. Each table and figure is introduced by the main-paper table, figure or section it supports.

\section{Experimental Setup}
\label{sec:appendix-setup}

\subsection{Architecture and Comparison With MuViT}
\label{sec:appendix-muvit}

\epssegformer{} builds on the encoder and the per-level decoders of MuViT (cited in the main paper). \Cref{tab:muvit_diff} lists what the two share and every difference we know of between our implementation and the official one, from the encoder to the training recipe. It expands the one-sentence summary of the changes in Section~\mainref{sec:method} of the main paper. The most consequential differences are in the masking (the center token of every level is always masked, dropped tokens are allocated uniformly over the levels, and the drop ratio follows a stepped schedule) and in the task (a joint classification and reconstruction objective instead of pre-training).

\input{fig/tab_muvit_diff}

The two decoder variants of Section~\mainref{sec:training} of the main paper are described next.

\paragraph{Decoders of MuViT.}
Each of the three decoders has two transformer layers of width 192 with one attention head and an MLP ratio of 2, preceded by a linear layer, a GELU and layer normalization on the queries and the context; it has 636,313 parameters, so the three decoders have 1.91M parameters together.

\paragraph{CNN Decoder of \muvitenc{}.}
The decoder reads the finest-view token map $z^{(1)}$. A transposed convolution with kernel and stride $P$ maps it to 256 channels, followed by batch normalization and a GELU; a $3\times3$ convolution with 256 channels and a GELU follows, and a $1\times1$ convolution gives one channel. The decoder has 1.82M parameters. The transposed convolution does not overlap (kernel and stride $P$), so the $3\times3$ convolution is the only mixing between patches. A dropped patch is therefore reconstructed from the mask token and a one-pixel margin of its neighbors, and for $P=5$ the center pixel $p$ lies in the inner $3\times3$ block of the dropped center patch and is reconstructed from the mask token alone. The decoder cannot infer a dropped patch from its wider context (Section~\mainref{sec:training} of the main paper).

\subsection{Dataset Composition}
\label{sec:appendix-data}

\Cref{tab:data} lists the BetaSeg cells, their class frequencies and the simulated labels. It holds the numbers behind the data description in Section~\mainref{sec:data} of the main paper: the class shares, the 10,029 training labels, the 85/15 split into training and validation slices, and the evaluation slice.

The evaluation slice~627 has index 626 when slices are counted from 0, as in our code. The validation slices carry labels drawn by the same sampler at twice the per-class density (3,596 labels, or 7.8 per slice against 3.8 for training).

Label sampling: from each training slice the sampler draws one pixel for each class present and two for Nucleus, which 1,225 of the 2,611 training slices (46.9\%) contain. Nucleus therefore receives 2,450 labels, 0.94 per slice on average, close to the one per slice of the other classes, which gives the class shares of \cref{tab:data}. Classes absent from a slice get no label.

Intensities are z-scored with the mean (129.989) and standard deviation (SD, 42.243) of c1--c3, also for high\_c4.

\input{fig/tab_data}

\subsection{Training and Augmentation}
\label{sec:appendix-impl}

The following paragraphs give the training schedule, the augmentation parameters and geometry, the padding fractions and the inference cost to which Section~\mainref{sec:impl} of the main paper refers.

\paragraph{Optimization Schedule.}
We use AdamW with weight decay $10^{-4}$. An epoch is 100 iterations of batch size 96, or 9,600 samples. Validation views are not augmented. The drop ratio $\rho$ starts at 0.70 and falls by 0.15 every 500 iterations until it reaches 0.25 after 1,500 iterations (15 epochs). The three-seed runs use early stopping on the validation loss with a patience of 40 epochs, counted from epoch 60 on, and at most 300 epochs; the ablations run for a fixed 200 epochs.

\paragraph{Augmentation Parameters.}
\Cref{tab:aug_params} lists the probabilities and parameters of the augmentation used in all augmented runs. It is the table to which Section~\mainref{sec:impl} of the main paper points for the augmentation parameters.

\input{fig/tab_aug_params}

\paragraph{Augmentation Geometry.}
The geometric transformation (flips, quarter turns, a small rotation and a smooth elastic field) acts on the sampling positions $p+s_\ell i$ of Eq.~(\mainref{eq:views}) of the main paper, and $I$ is sampled bilinearly at the transformed positions. The token coordinates of Section~\mainref{sec:views} remain those of the nominal grid. In view $\ell$, the blur widths and noise standard deviations are divided by the stride $s_\ell$, so the blur has the same physical width in every view (Section~\mainref{sec:training} of the main paper).

\definecolor{sgError}{HTML}{D946EF}
\providecommand{\qgchip}[1]{{\setlength{\fboxsep}{0pt}\setlength{\fboxrule}{0.3pt}\fcolorbox{black!70}{#1}{\rule{0pt}{0.62em}\rule{0.62em}{0pt}}}}
\newlength{\sfw}\setlength{\sfw}{\dimexpr(\textwidth-3pt)/2\relax}%
\newcommand{\sfpanel}[3][]{%
  \begin{tikzpicture}[inner sep=0pt,outer sep=0pt]
    \node[anchor=north west] (sfimg) at (0,0) {\includegraphics[width=\sfw]{figs/supp/#3.png}};
    \node[anchor=north west,fill=black,fill opacity=0.65,text opacity=1,text=white,rounded corners=1.5pt,inner xsep=3pt,inner ysep=2pt,font=\qgoverlayfont] at ([shift={(3pt,-3pt)}]sfimg.north west) {#2};
    #1
  \end{tikzpicture}}
\newlength{\sfh}\setlength{\sfh}{\dimexpr\sfw*5037/10000\relax}
\newlength{\sfbarh}\setlength{\sfbarh}{\dimexpr\sfh-31pt\relax}
\newcommand{\sfcbar}{%
  \begin{scope}[shift={(sfimg.north west)}]
    \fill[black,fill opacity=0.65,rounded corners=1.5pt] (3pt,-18pt) rectangle (31pt,-\sfh+3pt);
    \node[anchor=north west] at (7pt,-23pt) {\includegraphics[width=6pt,height=\sfbarh]{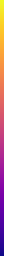}};
    \draw[white,line width=0.3pt] (7pt,-23pt) rectangle (13pt,-23pt-\sfbarh);
    \foreach \v in {1.00,0.75,0.50,0.25}{%
      \pgfmathsetlength{\sfy}{-23pt-(1-\v)/0.75*\sfbarh}%
      \draw[white,line width=0.3pt] (13pt,\sfy) -- (15pt,\sfy);
      \node[anchor=west,text=white,font=\sffamily\bfseries\fontsize{6}{6}\selectfont] at (16.5pt,\sfy) {\v};}
  \end{scope}}
\newlength{\sfy}
\newcommand{\sfrow}[2]{\par\nointerlineskip\hbox to\textwidth{#1\hfil#2}\nointerlineskip\vskip3pt}
\newcommand{\sflast}{}
\newcommand{\sflegsep}{\hspace{1.2em}}
\newcommand{\sflegend}[1]{\vskip2pt\par\begingroup\centering\small #1\par\endgroup}

\begin{figure*}[!t]
  \centering
  \includegraphics[width=\textwidth]{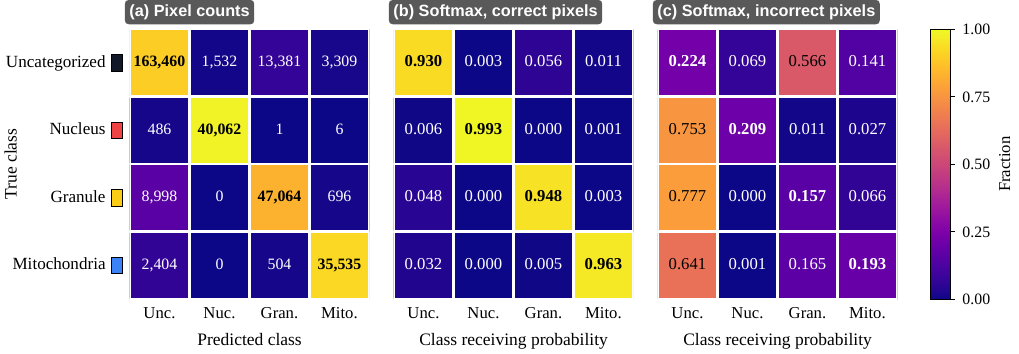}
  \caption{\textbf{Before Post-Processing, the Errors Are Concentrated Between Granules, Mitochondria and the \mbox{Uncategorized} Class, and Wrong Predictions Are Less Confident Than Correct Ones.}
  TTA3 prediction without post-processing (seed~74) on cell~high\_c4, slice 627 (317{,}438 in-cell pixels); rows are the true class.
  \textbf{(a)}~Confusion matrix: pixel counts, shaded by the share of the true class, so the diagonal is the per-class recall (0.900, 0.988, 0.829, 0.924).
  \textbf{(b)}~Mean softmax output over the correctly classified pixels, by true class; the diagonal, the probability given to the true class, is 0.93--0.99.
  \textbf{(c)}~The same over the misclassified pixels: the largest entry of a row is the wrong class that was chosen (0.753 for nucleus pixels called uncategorized), and the diagonal (0.16--0.22) is the probability left to the true class, still the second highest in every row.
  One plasma color bar serves all panels; diagonals are bold; the squares left of (a) are the class colors of \cref{fig:supp_tta3_postproc}.\looseness=-1}
  \label{fig:supp_tta3_matrices}
\end{figure*}
\begin{figure*}[t]
  \centering
  \sfrow{\sfpanel{(a) Misclassified pixels: 31{,}317 (9.87\%)}{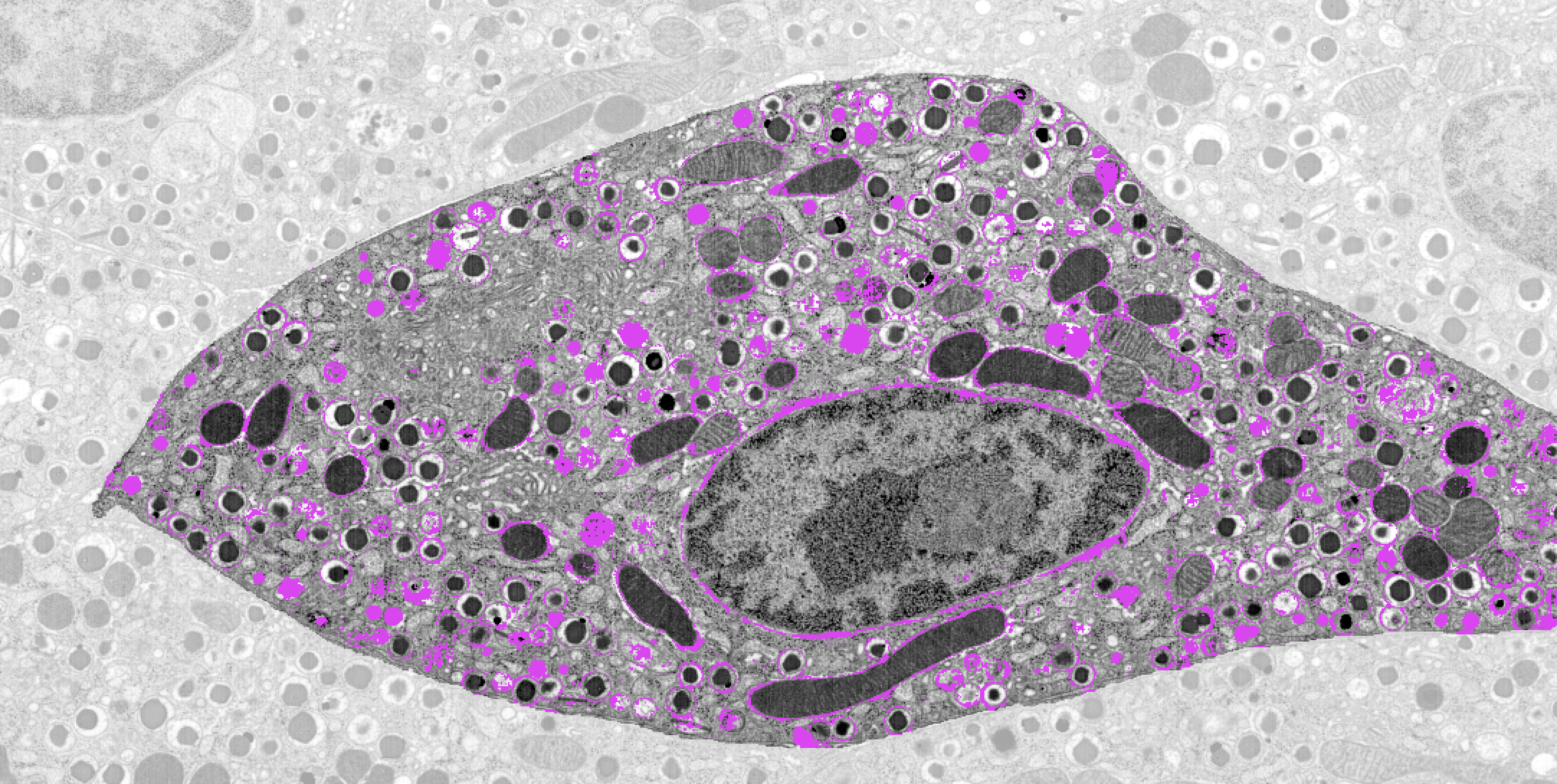}}{\sfpanel{(b) True class of the misclassified pixels}{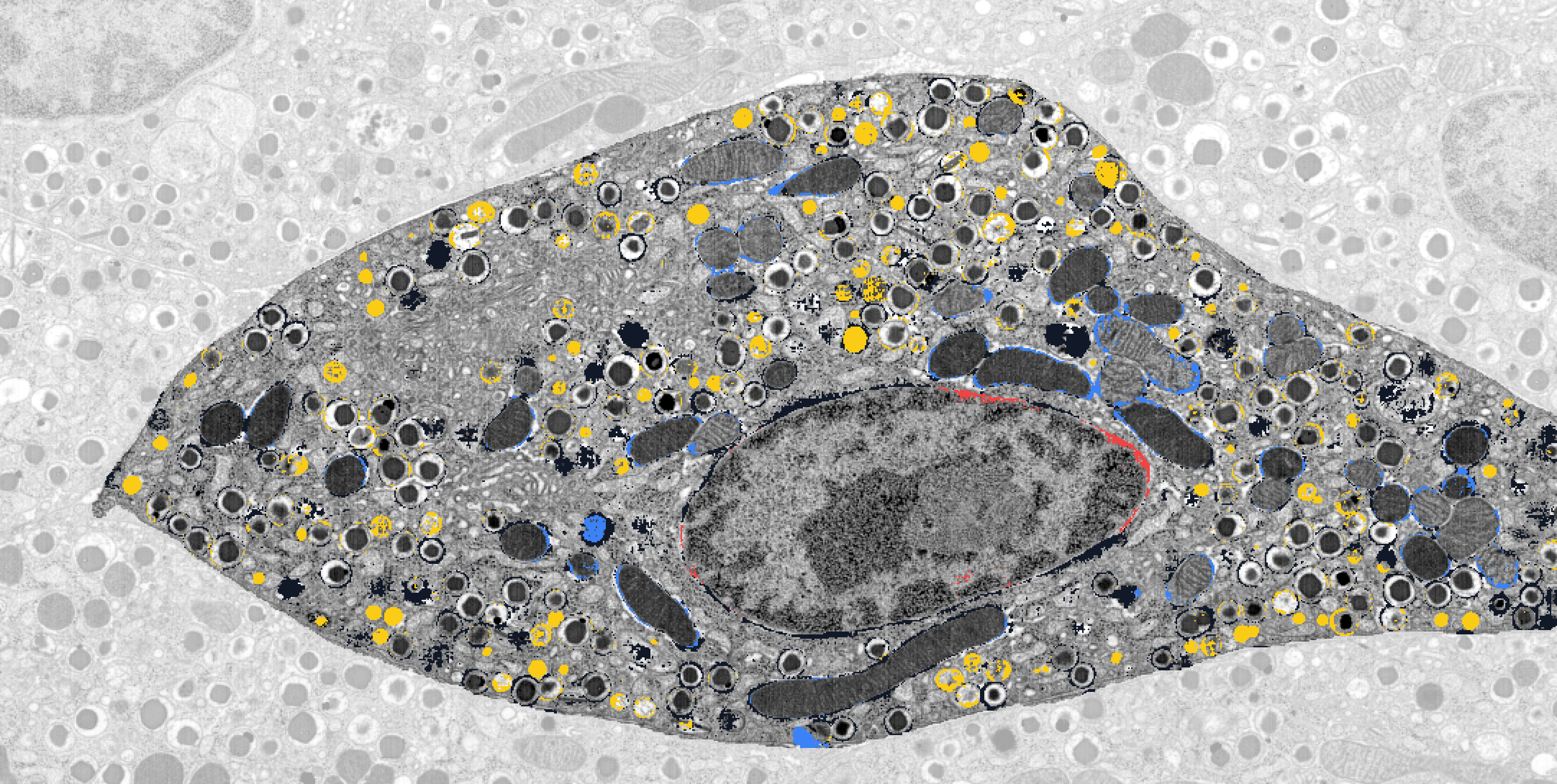}}
  \sfrow{\sfpanel[\sfcbar]{(c) Softmax of the wrong prediction}{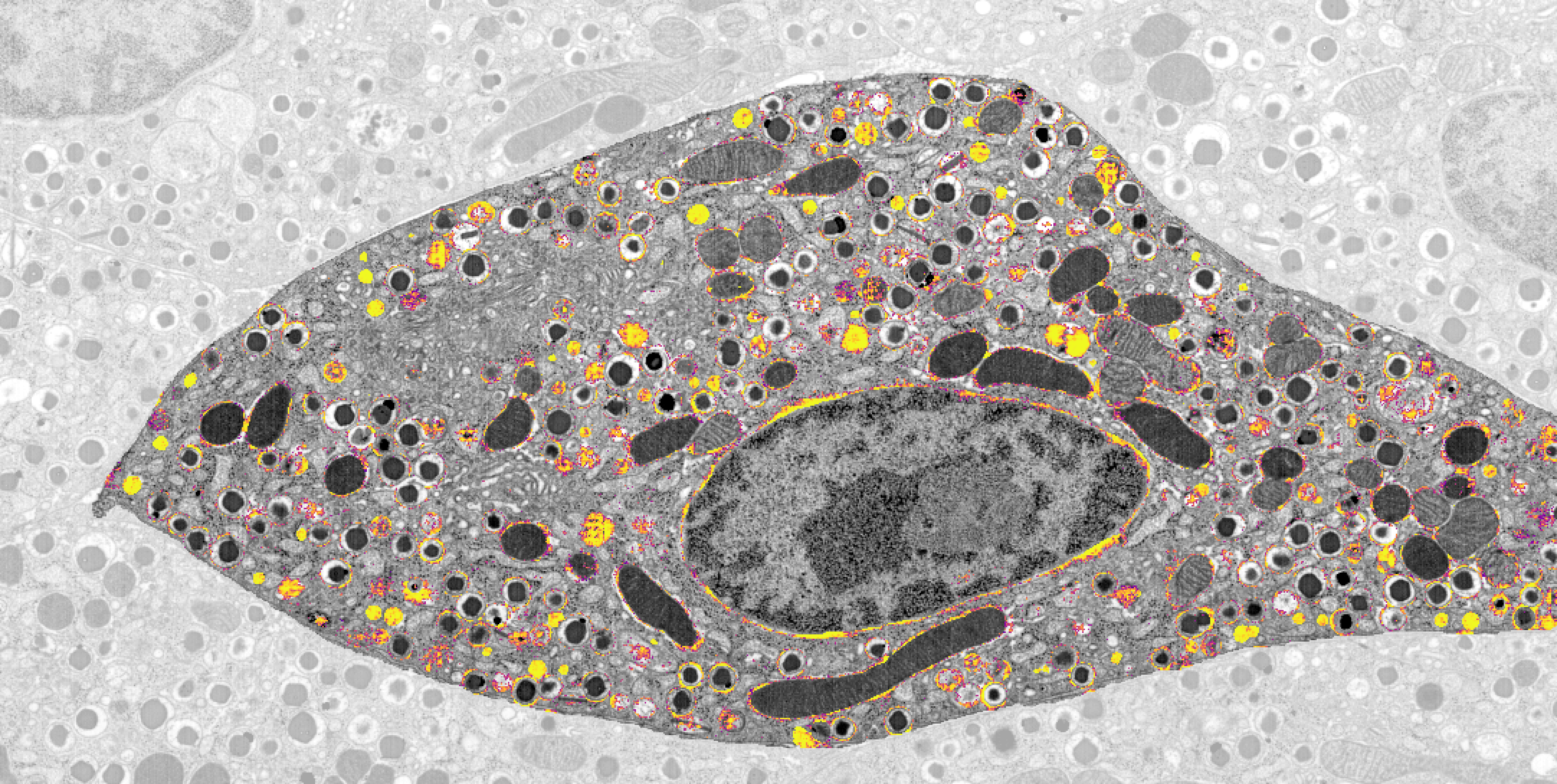}}{\sfpanel{(d) Predicted class of the misclassified pixels}{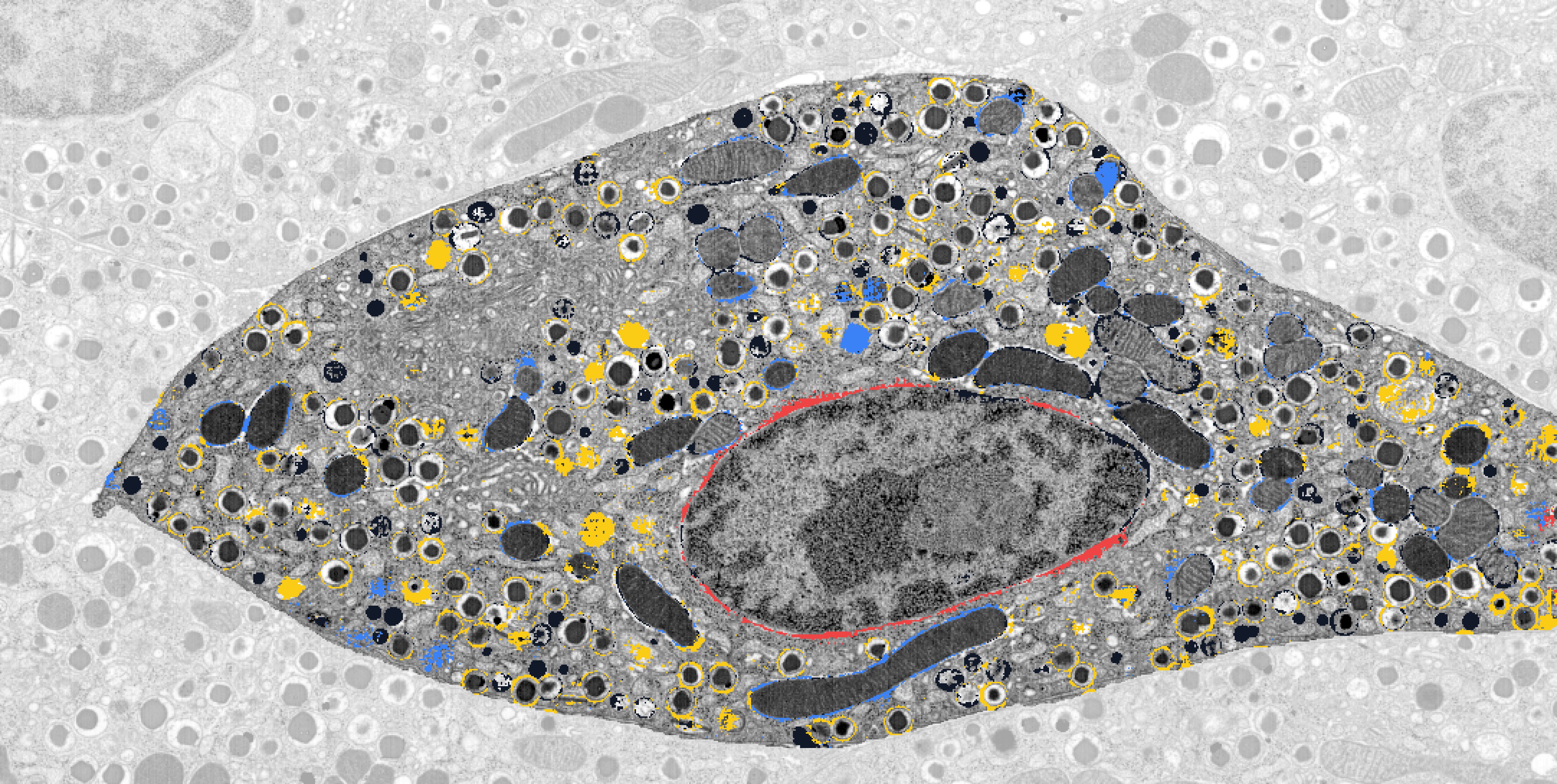}}\sflast
  \sflegend{\qgchip{qgOther}~uncategorized\sflegsep \qgchip{qgNucleus}~nucleus\sflegsep \qgchip{qgGranules}~granule\sflegsep \qgchip{qgMito}~mitochondria~(b, d)\sflegsep \qgchip{sgError}~misclassified pixel (a)}
  \caption{\textbf{Before Post-Processing, Nine in Ten Errors Are Confusions of Granules and Mitochondria With the \mbox{Uncategorized} Class, and a Third of the Wrong Pixels Are Predicted With Confidence Above 0.9.}
  Misclassified pixels of the TTA3 prediction without post-processing of \cref{fig:supp_tta3_matrices}; only these pixels are colored. Each pixel is a $4{\times}4$ block, without interpolation; the EM image is contrast-stretched for display and faded (35\% contrast) outside the cell mask.
  \textbf{(a)}~Misclassified pixels (magenta), 9.87\% of the 317{,}438 in-cell pixels; panel (e) of \cref{fig:supp_tta3_postproc}.
  \textbf{(b)}~Colored by the true class: 58.2\% uncategorized, 31.0\% granules, 9.3\% mitochondria, 1.6\% nucleus.
  \textbf{(c)}~Colored by the softmax probability of the predicted class (scale 0.25--1, the lowest value a four-class prediction can have): mean 0.784, 34.9\% above 0.9, 1.8\% below 0.5.
  \textbf{(d)}~Colored by the predicted class: 44.3\% granules, 38.0\% uncategorized, 12.8\% mitochondria, 4.9\% nucleus.
  Of all errors, 89.7\% are granules or mitochondria called uncategorized or the reverse (counts in \cref{fig:supp_tta3_matrices}(a)). The few nucleus errors form a thin rim along part of the nuclear envelope.}
  \label{fig:supp_tta3_errors}
\end{figure*}
\begin{figure*}[t]
  \centering
  \sfrow{\sfpanel{(a) EM image}{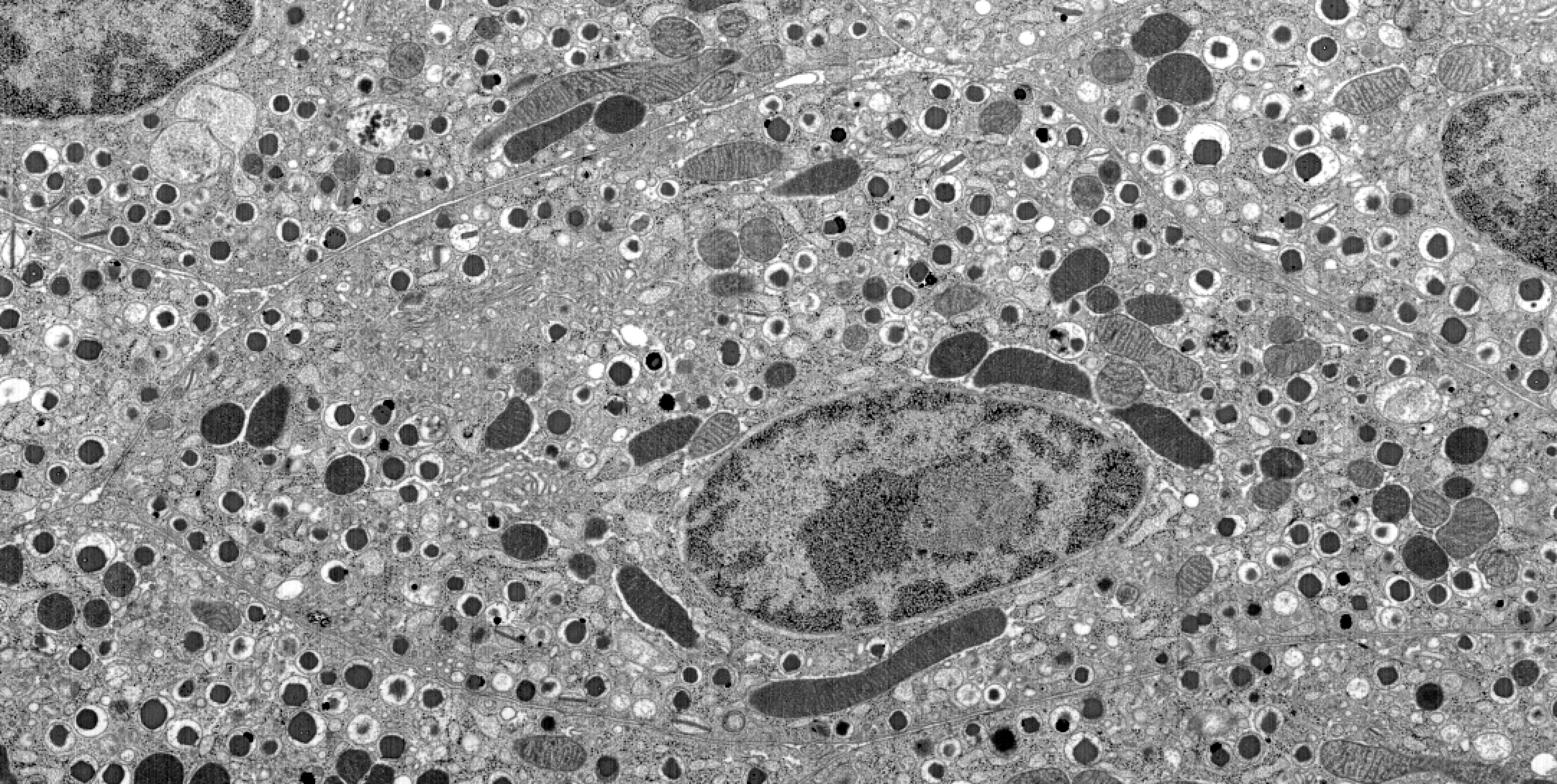}}{\sfpanel{(b) Ground truth}{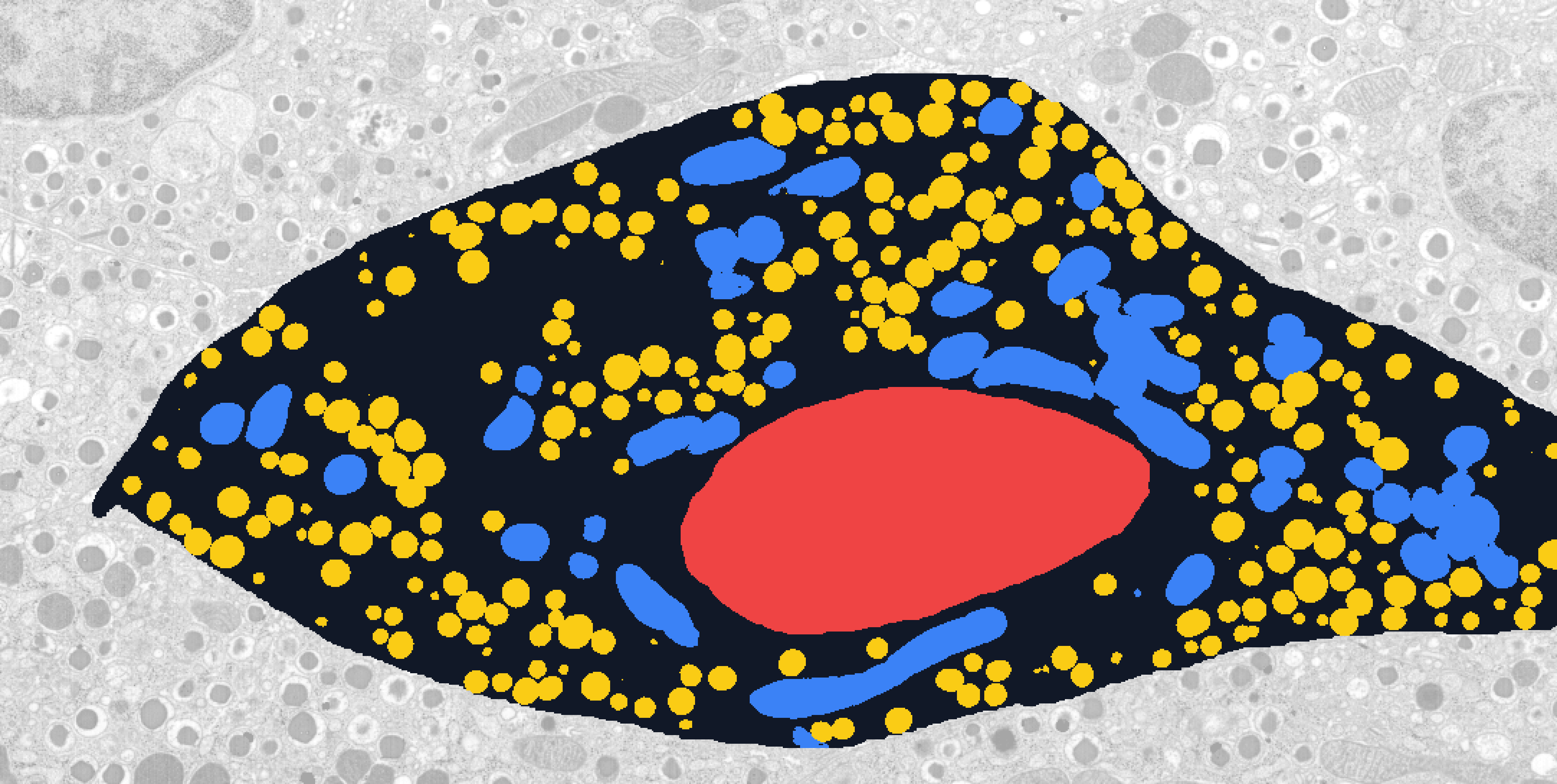}}
  \sfrow{\sfpanel{(c) Before post-processing (Avg.\ Dice 0.9005)}{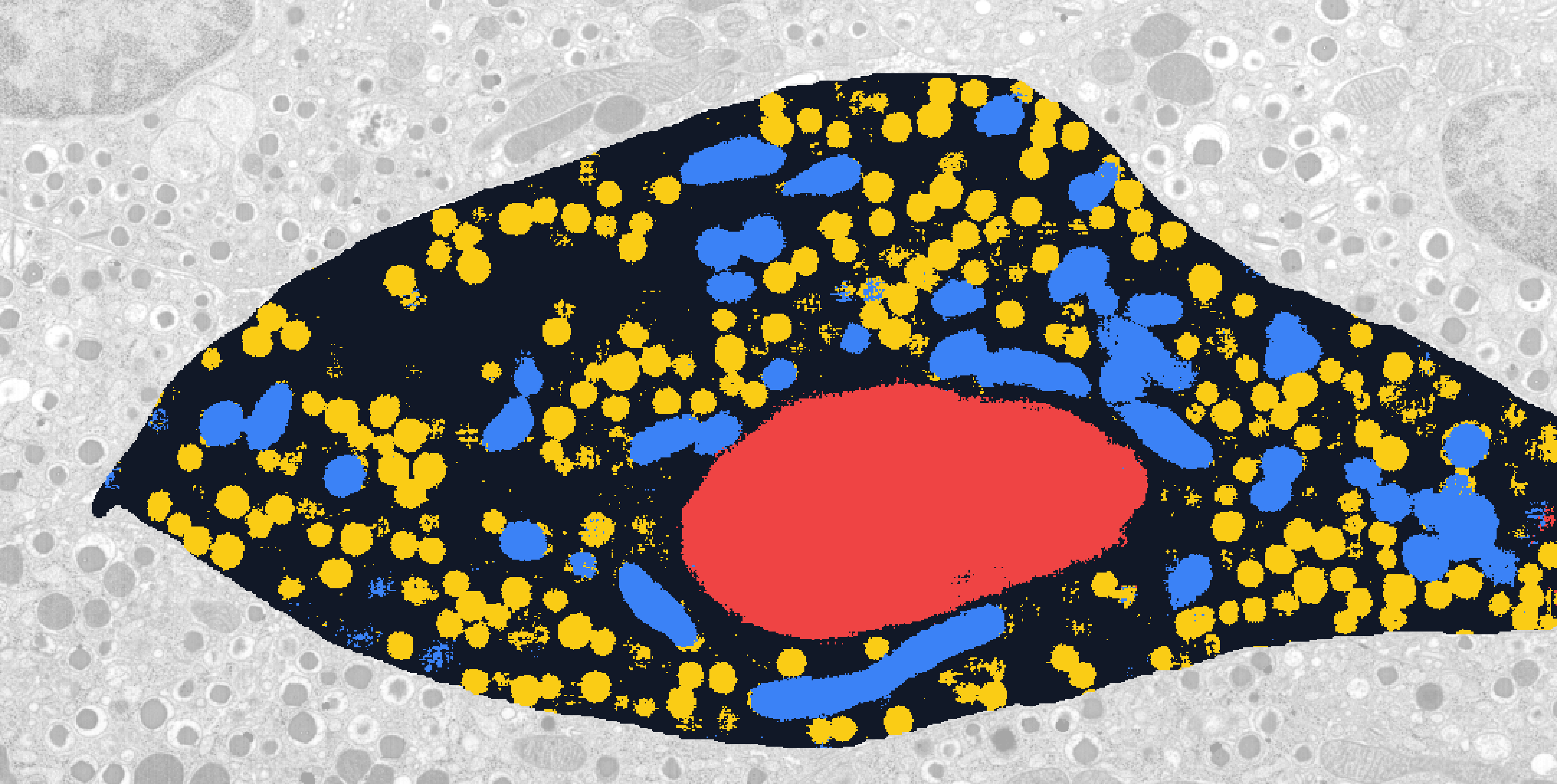}}{\sfpanel{(d) After post-processing (Avg.\ Dice 0.9124)}{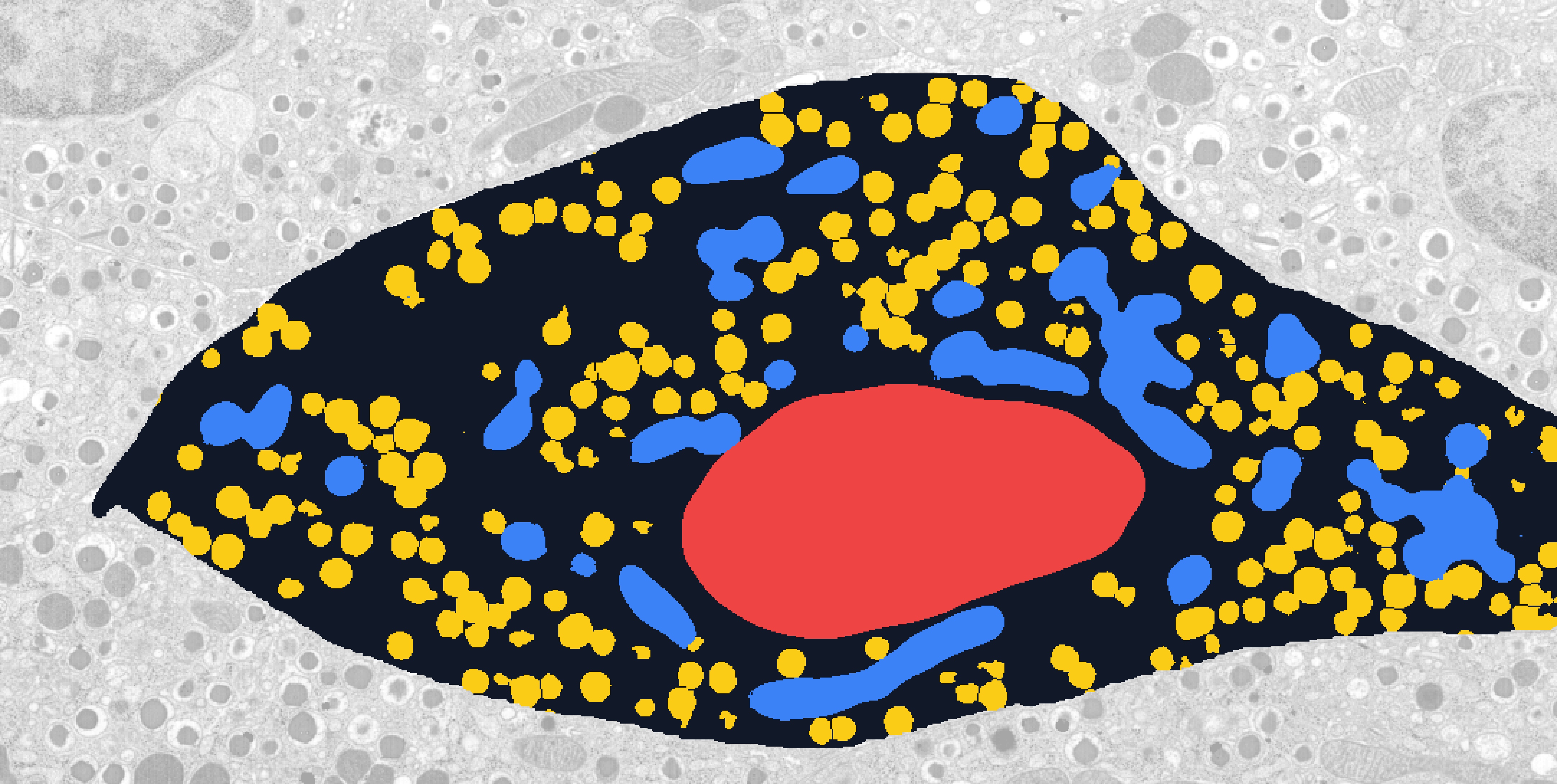}}
  \sfrow{\sfpanel{(e) Errors before: 31{,}317 px (9.87\%)}{s2/err_raw}}{\sfpanel{(f) Errors after: 27{,}344 px (8.61\%)}{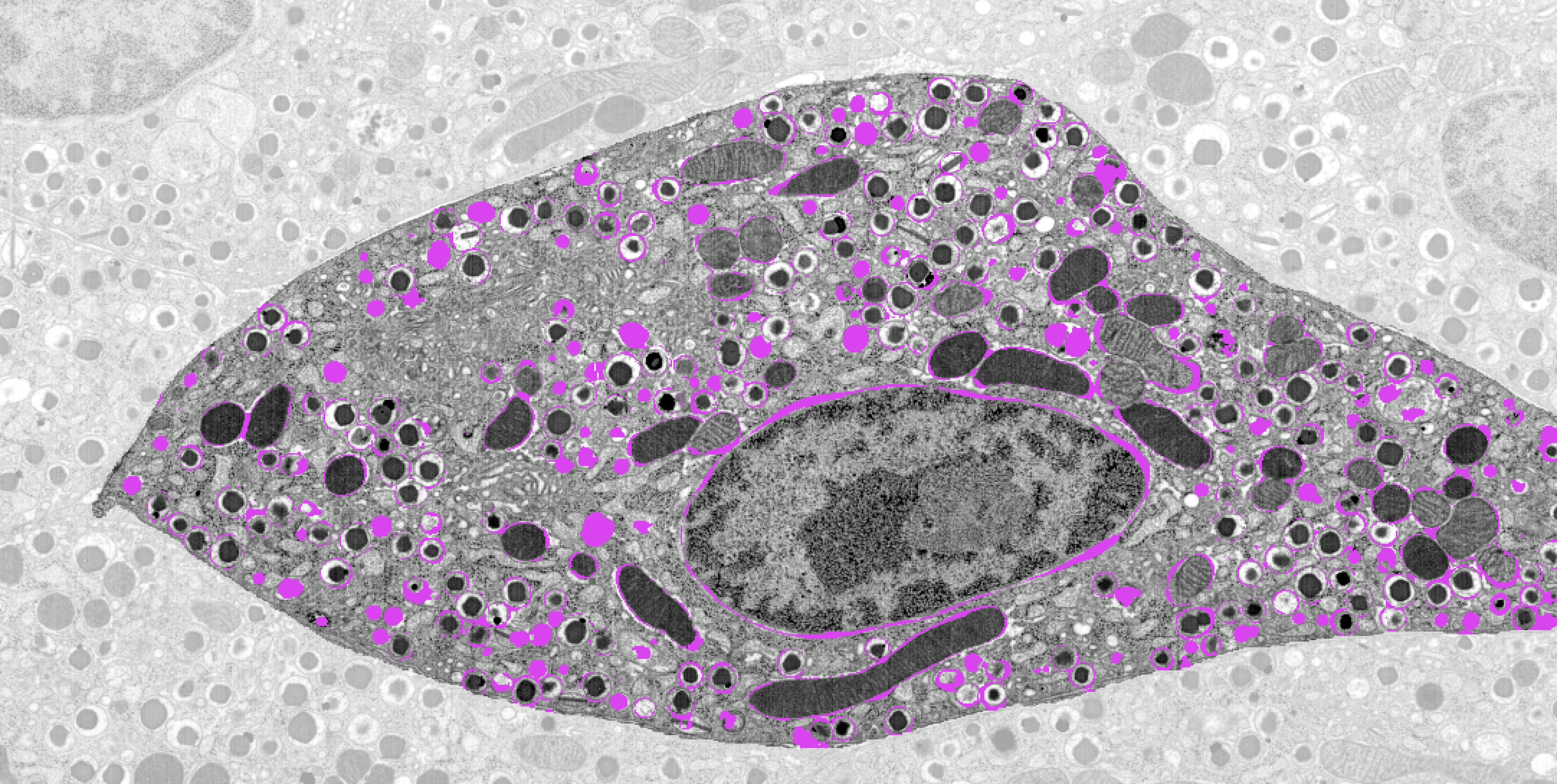}}\sflast
  \sflegend{\qgchip{qgOther}~uncategorized\sflegsep \qgchip{qgNucleus}~nucleus\sflegsep \qgchip{qgGranules}~granule\sflegsep \qgchip{qgMito}~mitochondria\sflegsep \qgchip{sgError}~misclassified pixel (e, f)}
  \caption{\textbf{Post-Processing Removes Speckle and Separates Fused Granules, and Lowers the Share of Misclassified Pixels From 9.87\% to 8.61\%.}
  Same slice and prediction as \cref{fig:supp_tta3_matrices,fig:supp_tta3_errors}, drawn as in \cref{fig:supp_tta3_errors} in the class colors of Figure~\mainref{fig:qual_grid}; the EM image in (a) is at full contrast. The pipeline is described in Section~\mainref{sec:inference}.
  \textbf{(c)}~TTA3 prediction without post-processing (Avg.\ Dice 0.9005): isolated errors appear as speckle inside objects, and granules are frayed or merged.
  \textbf{(d)}~After post-processing (Avg.\ Dice 0.9124; pixel accuracy 90.13\%\,$\rightarrow$\,91.39\%).
  \textbf{(e,\,f)}~Misclassified pixels (magenta) before and after: 31{,}317 (9.87\%) and 27{,}344 (8.61\%) of the 317{,}438 in-cell pixels. Post-processing fixes 8{,}708 wrong pixels and breaks 4{,}735 correct ones, a net gain of 3{,}973. Most remaining errors are thin rings around object boundaries.}
  \label{fig:supp_tta3_postproc}
\end{figure*}

\paragraph{Border Padding.}
With $85{\times}85$ views at scales $\{1,4,8\}$, the mean padded fraction of a view is 0.02\%, 4.7\% and 25.0\% for the 10,029 training labels, and 0.6\%, 9.2\% and 32.3\% for the in-cell pixels of slice~627. The coarsest view contains padding for 99.7\% of the training labels and for every evaluation pixel.

\paragraph{Inference Cost.}
We timed the existing prediction path on slice~627 of high\_c4 (317,438 in-cell pixels, 2,560 batches of 124 pixels) with each of the three final checkpoints, on one Tesla V100-SXM2 (32\,GB) with PyTorch 2.8.0, CUDA 12.8, fp32 and 3 CPU cores. The time covers coordinate collection, CPU patch sampling, transfers, the forward pass (including the decoders) and the CPU assembly of the probabilities and of the TTA3 average; it excludes checkpoint loading (about 28\,s), reading the volume, warm-up, scoring and post-processing. Without TTA3, a slice takes $5{,}812\pm72$\,s (mean $\pm$ SD of the three seeds; 54--55 pixels/s), or 1.6\,h. TTA3 runs the three views one after the other and takes $17{,}466\pm72$\,s (18 pixels/s), or 4.9\,h, three times as long. Peak GPU memory is 2.18\,GiB allocated and 2.99\,GiB reserved by PyTorch (excluding the CUDA context), the same with and without TTA3.

Extrapolated by pixel count at the throughput of this slice, the 193.7 million in-cell voxels of high\_c4 (610 times the slice) take about 41 GPU-days per seed, or 123 with TTA3; the whole 1022\,$\times$\,545\,$\times$\,1082 volume (1,899 times the slice) would take 128 and 384 GPU-days. These are serial single-GPU times for one seed; the pixels are independent, so the work splits across GPUs. The estimate assumes the throughput of slice~627 holds for the other slices; the share of padded pixels and the per-slice overhead can change it. Part of the time may be CPU-bound, as the job had only 3 cores and the patch sampling runs on the CPU; we did not profile GPU utilization.

\section{Results of the Final Model}
\label{sec:appendix-final}

This part unpacks the headline result of the main paper (Table~\mainref{tab:comparison}, Figure~\mainref{fig:qual_grid} and Section~\mainref{sec:res-compare}): the per-class Dice of every seed, and the confusions, the errors and the effect of post-processing for the run of Figure~\mainref{fig:qual_grid} (\cref{fig:supp_tta3_matrices,fig:supp_tta3_errors,fig:supp_tta3_postproc}).

\subsection{Per-Class Results of the Final Model}
\label{sec:appendix-perclass}

\input{fig/tab_perclass}

\Cref{tab:perclass} gives the per-class Dice of the final model for each seed with TTA3, with and without post-processing. The TTA3 and post-processing mean is the \epssegformer{} row of Table~\mainref{tab:comparison} (the $0.9090\pm0.0048$ of Section~\mainref{sec:res-compare}, with $0.8977\pm0.0056$ for TTA3 alone), and seed~74 is the run of Figure~\mainref{fig:qual_grid} of the main paper. The raw row of Table~\mainref{tab:comparison} (single forward pass, $0.8903\pm0.0073$) is the Raw column of Table~\mainref{tab:config_selection} and \cref{tab:config_full}.

\subsection{Predictions on the Full Slice}
\label{sec:appendix-fullslice}

\Cref{fig:supp_tta3_matrices,fig:supp_tta3_errors,fig:supp_tta3_postproc} show the run of Figure~\mainref{fig:qual_grid} of the main paper on the whole slice: the evaluation of one trained model (seed~74) on the full slice 627 of cell~high\_c4, using test-time augmentation with three views (TTA3: the slice as is, flipped horizontally and flipped vertically; the class probabilities of the three views are averaged per pixel).
Only the 317{,}438 in-cell pixels are evaluated.
Before post-processing, the pixel accuracy is 90.13\% and the Avg.\ Dice is 0.9005.

\Cref{fig:supp_tta3_matrices} shows which classes are confused behind the per-class Dice of \cref{tab:perclass} and Section~\mainref{sec:res-compare}, where Granule scores lowest, and how confident the predictions are.

\Cref{fig:supp_tta3_errors} locates the misclassified pixels of the same prediction and gives their true and predicted classes and their confidence. Section~\mainref{sec:res-compare} reports how close the misclassified pixels of the TTA3 output without post-processing lie to a class boundary; this figure shows where they are and which classes they confuse.

\Cref{fig:supp_tta3_postproc} compares the prediction before and after post-processing, which Section~\mainref{sec:inference} describes and Section~\mainref{sec:res-final} evaluates. Its panel (d) is the class map of the \epssegformer{} panel in Figure~\mainref{fig:qual_grid}, the seed-74 run behind the TTA3 + Post row of Table~\mainref{tab:comparison}.

\subsection{Tuning Protocol of the Post-Processing}
\label{sec:appendix-postproc}

We tuned the post-processing (Section~\mainref{sec:inference} of the main paper) on the evaluation slice, so its gain on the whole slice would be optimistic. To limit this, we split the slice at column 541 into a left half (140,785 in-cell pixels) and a right half (176,653), ranked candidate pipelines on the left half with four earlier models and without TTA3, and fixed the pipeline before the final model was evaluated and before any TTA3 output existed. We report the change in Avg.\ Dice on the right half (Section~\mainref{sec:res-final}), which played no part in the ranking. It holds 56\% of the in-cell pixels but 83\% of the Nucleus pixels, so its change need not equal that of the whole slice. The split does not cover the Gaussian widths, which we set from ground-truth object sizes in the whole slice, or the absolute post-processed Avg.\ Dice.

In the pipeline of Section~\mainref{sec:inference} of the main paper, each class in turn (Nucleus, Mitochondria, Granule) takes every undecided pixel where its probability, smoothed with a class-specific Gaussian normalized by the cell mask, exceeds the summed smoothed probability of the classes not yet decided; Uncategorized takes the rest. After the watershed, pixels on its cuts and in components below the minimum size take their next most probable class. The pipeline uses Gaussian widths $\sigma=8$, $4$ and $1.5$\,px for Nucleus, Mitochondria and Granule (in the order in which the classes are decided), a watershed with a minimum peak distance of 4\,px, and a minimum connected-component size of 50 pixels.

\section{Ablations of the Design Choices}
\label{sec:appendix-ablations}

Except for the tables that list three seeds, the tables in this part are single training runs (Section~\mainref{sec:seeds}), and as in the main paper we do not read a difference below 0.009, the single-run threshold of Section~\ref{sec:appendix-threshold}, as real. Unless a table says otherwise, the runs use the scales $\{1,2,4\}$ and drop only the center token of the finest view (Section~\mainref{sec:configs}). The order is: the choice to drop the center token, which the single-view ViT of \cref{sec:appendix-summary} tests; the two steps of Table~\mainref{tab:roadmap}; the comparison with a single scale at the final configuration (\cref{tab:scales1_full}); the crop scales and the kernel (\cref{tab:scales_full}); and the view size, whose spread leads to the three-seed runs of Part~\ref{sec:appendix-selection}.

\subsection{Dropping the Center Token in a Single-View ViT}
\label{sec:appendix-summary}

\Cref{tab:masking-summary} compares four ways of dropping tokens in a single-resolution ViT with the CNN decoder (one view). These runs led us to drop the center token in all other experiments. They are the evidence for this choice that Section~\mainref{sec:limitations} of the main paper refers to (see also Section~\mainref{sec:hidden}).

\begin{table}[!t]
\centering
\caption{\textbf{Masking~4 (M4) Gives the Highest Mean Avg.\ Dice, With Small Margins.}
Single-view ViT with the CNN decoder, compared across four drop patterns at a 25\% drop ratio. Each value is averaged over three reconstruction losses (inpainting: dropped tokens only; visible tokens only; all tokens).
M1 drops tokens at random among those the segmentation head (S.H.) does not read, and the center token stays.
M2 is M1 with the center token always dropped.
M3 drops tokens at random among all tokens but the center, which stays.
M4 is M3 with the center token always dropped.
$\Delta$ is the effect of always dropping the center token (M2 minus M1, M4 minus M3); it is positive on average, but not in every setting. M4 leads M2 by 0.0019. Bold marks the best pattern per row. All runs are single runs.}
\label{tab:masking-summary}
\tabstylenarrow\setlength{\tabcolsep}{1.5pt}
\begin{tabular*}{\columnwidth}{@{\extracolsep{\fill}}lcccc@{\hspace{5pt}}ccc}
\toprule
\makecell[l]{View / patch\\size (px)} & \makecell[c]{S.H.\\kernel} & M1 & M2 & $\Delta$ & M3 & M4 & $\Delta$ \\
\midrule
$25{\times}25$ / $5{\times}5$ & $3{\times}3$ & 0.7114 & 0.7479 & $+0.0365$ & \textbf{0.7502} & 0.7334 & \dn{0.0168} \\
\cmidrule(lr){1-8}
\multirow{2}{*}{$27{\times}27$ / $3{\times}3$} & $3{\times}3$ & 0.7352 & 0.7503 & $+0.0151$ & \textbf{0.7642} & 0.7622 & \dn{0.0020} \\
 & $5{\times}5$ & 0.7333 & 0.7446 & $+0.0113$ & \textbf{0.7600} & \textbf{0.7600} & \phantom{$+$}$0.0000$ \\
\cmidrule(lr){1-8}
\multirow{2}{*}{$63{\times}63$ / $3{\times}3$} & $3{\times}3$ & \textbf{0.7688} & 0.7626 & \dn{0.0062} & 0.7470 & 0.7483 & $+0.0013$ \\
 & $5{\times}5$ & 0.7437 & 0.7733 & $+0.0296$ & 0.7734 & \textbf{0.7758} & $+0.0024$ \\
\cmidrule(lr){1-8}
$65{\times}65$ / $5{\times}5$ & $3{\times}3$ & 0.7860 & 0.7843 & \dn{0.0017} & 0.7584 & \textbf{0.7946} & $+0.0362$ \\
\midrule
\multicolumn{2}{l}{Mean} & 0.7464 & 0.7605 & $+0.0141$ & 0.7589 & \textbf{0.7624} & $+0.0035$ \\
\bottomrule
\end{tabular*}
\end{table}

\subsection{Center Dropping in a Single Run}
\label{sec:appendix-center}

This is the closest comparison we have for dropping the center token of every view, which \cref{tab:masking-summary} does not cover (it has a single view). The ablations drop only the center token of the finest view; the final model drops that of every view. For MuViT with $85{\times}85$ views and a $5{\times}5$ kernel, the single run with finest-view dropping (200 fixed epochs, seed~42; \cref{tab:viewsize_full}) scores 0.8882, against $0.8903\pm0.0073$ for the final runs over three seeds with early stopping and all-view dropping. The runs differ in the number of seeds, the training length and the dropping, so we attribute no difference to the dropping (Section~\mainref{sec:limitations} of the main paper).

\subsection{Hiding the Center Token at the Final Configuration}
\label{sec:appendix-center-final}

\Cref{tab:center_full} gives the direct test at the final configuration: the same model and three seeds, with the center token of every view kept visible in training. Section~\mainref{sec:res-scales} of the main paper summarizes it.

\input{fig/tab_center_full}

\subsection{Augmentation and Border Padding}
\label{sec:appendix-aug-pad}

\Cref{tab:aug_ctx} gives the per-setting results behind the first step of Table~\mainref{tab:roadmap} of the main paper (Section~\mainref{sec:res-augpad}): adding data augmentation.

\input{fig/tab_aug_ctx}

\Cref{tab:padding} gives the per-setting results behind the second step of Table~\mainref{tab:roadmap}: adding border padding to the augmented runs.

\input{fig/tab_padding}

\Cref{tab:padding_pixels} shows how the padded and unpadded runs of \cref{tab:padding} differ in their training and scored pixels. Padding adds 107 training labels at $27{\times}27$ and 1,664 (19.9\%) at $65{\times}65$, and the extra labels are mostly Uncategorized, Granule and Mitochondria. Scoring omits only a thin border strip, 0.6\% and 1.8\% of the in-cell pixels, so the published gains understate the gains on identical pixels by 0.002 to 0.003 on average. For \muvitenc{} at $65{\times}65$, the unpadded model scores the 30.5\% of pixels whose coarsest view is partly padded without having seen padded views in training, and most of the gain lies there. The table cannot separate this from the effect of the extra labels, since padded runs get both.

\input{fig/tab_padding_pixels}

\subsection{Single Scale Versus Multiple Scales at the Final Configuration}
\label{sec:appendix-scales1}

\Cref{tab:scales1_full} gives the per-seed results behind the paragraph on coarser views of Section~\mainref{sec:res-scales} of the main paper. The multi-view runs are the six rows of \cref{tab:config_full} with a $5{\times}5$ S.H.\ kernel, trained at the scales $\{1,4,8\}$. The single-view runs use the same models, view sizes, kernel and seeds (54, 74, 313) with only the finest view (scale~1). The three views are higher in all 18 seed pairs, by 0.0123 to 0.0427, and in each of the six configurations the mean difference (0.0156 to 0.0357) is above the 0.005 threshold of Section~\mainref{sec:seeds}. The single view has a third of the tokens (289 against 867 at $85{\times}85$), so its encoder self-attention costs about a ninth and its MLP and projections about a third of those of the multi-view model. It spans an eighth of the width of the coarsest view; a full-resolution view of that width ($680{\times}680$ px at $85{\times}85$ views and $5{\times}5$ patches) would need $136^2=18{,}496$ tokens. The comparison therefore does not separate the effect of the wider context from that of the extra tokens.

A pairing with a closer token count is the $85{\times}85$ single view (289 tokens) against the $27{\times}27$ three-view model (243 tokens). With matching seeds, the three-view model is higher in all six pairs, by $+0.0061$ to $+0.0294$ (mean $+0.0168$ for MuViT and about $+0.019$ for \muvitenc{}). The two models differ in more than the scales. The $27{\times}27$ model has $3{\times}3$ patches and the $85{\times}85$ model $5{\times}5$ patches, which changes the pixels per token at full resolution (9 against 25) and the size of the dropped center token at the finest scale (3 against 5 px). The $85{\times}85$ single view also sees 85 px at full resolution, against 27 px. We therefore do not read this pairing as a token-matched test.

\input{fig/tab_scales1_full}

\subsection{Crop Scales and S.H.\ Kernel Size}
\label{sec:appendix-scales}

\input{fig/tab_scales_full}
\input{fig/tab_viewsize_full}

\Cref{tab:scales_full} lists the scales $\{1,2,4\}$ and $\{1,4,8\}$ for every S.H.\ kernel size and both view sizes of the single-run ablation. These are the runs behind the scale and kernel paragraphs of Section~\mainref{sec:res-scales} of the main paper, grouped here by architecture. Replacing the scales $\{1,2,4\}$ by $\{1,4,8\}$ changes Avg.\ Dice by \dn{0.0056} to $+0.0110$ and by $+0.0028$ on average; seven of the eight changes are positive and five are below 0.005. With the scales $\{1,2,4\}$, the best S.H.\ kernel is $5{\times}5$ in three of the four combinations of view size and architecture, and $3{\times}3$ in the fourth, where it is $0.0032$ above $5{\times}5$. The larger kernels $9{\times}9$ and $13{\times}13$ fall $0.02$ to $0.07$ below the best kernel. The kernel size sets the number of head parameters (0.8M more for $5{\times}5$ than for $3{\times}3$), so these runs do not isolate the context.

\subsection{View Size}
\label{sec:appendix-viewsize}

\Cref{tab:viewsize_full} gives the single-run results for the view sizes that the main paper compares: the choice of the view size listed under the ablations in Section~\mainref{sec:configs}, and the spans of 0.0070 to 0.0199 across views of $65{\times}65$ to $105{\times}105$ that Section~\mainref{sec:res-select} quotes to show that single runs cannot rank the view sizes. Part~\ref{sec:appendix-selection} repeats the comparison of the $65{\times}65$ and $85{\times}85$ views with three seeds.

\section{Selection of the Final Configuration}
\label{sec:appendix-selection}

The eight configurations of Table~\mainref{tab:config_selection} differ in architecture, view size and S.H.\ kernel size and are trained with three seeds. The final configuration ranks first in every column of that table, with margins below the 0.005 we read as real (Section~\mainref{sec:res-select}).

\subsection{Per-Seed Results of the Configurations}
\label{sec:appendix-config}

Table~\mainref{tab:config_selection} of the main paper (Section~\mainref{sec:res-select}) shows the means over three seeds. \Cref{tab:config_full} gives the per-seed results behind it, the selected epochs, the token counts and the parameter counts, and adds four $27{\times}27$ configurations from an earlier set that are not ranked.

\input{fig/tab_config_full}

\subsection{Pairwise Comparisons of the Configurations}
\label{sec:appendix-pairs}

Single runs cannot rank the view sizes: across views of $65{\times}65$ to $105{\times}105$, Avg.\ Dice spans 0.0070 to 0.0199, depending on architecture and kernel (\cref{tab:viewsize_full}), which is why Table~\mainref{tab:config_selection} uses three seeds.
The comparisons below pair runs with the same seed that differ in one setting; the pair counts are descriptive, because all runs share the label set and the evaluation slice.

\paragraph{Architecture.}
Three seeds do not separate MuViT from \muvitenc{} by accuracy.
Over the four matched configurations (12 pairs), \muvitenc{} is higher by 0.0003 raw and lower by 0.0030 post-processed, both below 0.005.
The raw sign depends on the view size: \muvitenc{} is higher by 0.0043 at $65{\times}65$ and lower by 0.0037 at $85{\times}85$. \muvitenc{} reaches its best checkpoint earlier, after 107 epochs on average against 170 for MuViT (earlier in 11 of 12 pairs; \cref{tab:config_full}); the validation losses of the two architectures differ (Section~\mainref{sec:ckpt}), so this shows when each loss stops improving, not how well a model fits.

\paragraph{View Size.}
Pooled over both architectures, the $65{\times}65$ and $85{\times}85$ views do not differ: the $65{\times}65$ view is higher by 0.0013 raw and 0.0018 post-processed, in 6 and 7 of 12 pairs.
The sign depends on the architecture.
For \muvitenc{}, the $65{\times}65$ view is higher by 0.0053 raw (in 5 of 6 pairs).
For MuViT, it is lower by 0.0027 (in 1 of 6), and the advantage of the $85{\times}85$ view is confined to the $5{\times}5$ kernel: $+0.0066$ with $5{\times}5$, but \dn{0.0012} with $3{\times}3$.

\paragraph{Kernel.}
The $5{\times}5$ kernel scores higher than $3{\times}3$ by 0.0055 raw and 0.0026 post-processed, in 9 and 8 of 12 pairs.
The raw difference is at the limit of what we read as real, so we take it as a mild preference.

\subsection{Reading Threshold}
\label{sec:appendix-threshold}

The seed SD of a configuration lies between 0.001 and 0.011, and the pooled SD is 0.0062 to 0.0067 per measure; we use 0.0065. The difference of two three-seed means then has a standard error of $0.0065\sqrt{2/3}\approx0.005$, and the difference of two single runs one of $0.0065\sqrt{2}\approx0.009$. We read a difference as real only above these values, so the thresholds are one standard error of the difference. Separating two configurations at about two standard errors would take about 0.011 for three-seed means and 0.018 for single runs. The single-run value assumes that the single-run ablations have the same seed SD as the three-seed runs; we have not measured this, because the ablations use one seed (42). Differences use unrounded per-seed values and can deviate by 0.0001 from those of the printed means.

\subsection{Analysis of Variance}
\label{sec:appendix-anova}

Section~\mainref{sec:res-select} reads the margins between the eight configurations against the 0.005 rule of Section~\mainref{sec:seeds}. A significance test, which the main paper does not report, is added here. A one-way analysis of variance of the eight configurations of Table~\mainref{tab:config_selection} (three runs each, 16 residual degrees of freedom, pooled within-configuration variance) does not reject equal means in any column: $F_{7,16}=1.20$, $1.26$, $0.75$ and $0.69$ ($p=0.36$, $0.33$, $0.64$ and $0.68$) for Raw, TTA3, Post and TTA3 + Post. The test treats the runs as independent and the variances as equal, and it ignores that all runs share the label set and the evaluation slice.

\subsection{Optimism From Selecting the Best of Eight}
\label{sec:appendix-optimism}

The final configuration is the best of eight. If the eight were equal, the best of eight three-seed means would exceed the true value by about 0.005 (0.0054 raw; pooled seed SD 0.0065), so the scores of the final configuration are optimistic by about that much. The search also covered the $27{\times}27$ configurations of \cref{tab:config_full} and the single-run ablations, so selection has a larger effect than this figure. The post-processing was tuned on the same slice, as Section~\mainref{sec:postproc} of the main paper describes.

\subsection{A Second Slice of the Held-Out Cell}
\label{sec:appendix-slice577}

The main paper scores every model on slice~627 of high\_c4, the slice of the published baselines, and selects the configuration and tunes the post-processing on it. To check that the results do not depend on this one slice, we scored the same 24 trained models (eight configurations, three seeds) on slice~577 of the same cell, 50 slices away, with the same metric, on the in-cell pixels of that slice (Section~\mainref{sec:configs}). Nothing was retrained or retuned: the checkpoints, TTA3 and the post-processing pipeline are those used for slice~627. \Cref{tab:slice577} gives the results next to those of slice~627.

\input{fig/tab_slice577}

\paragraph{Scores.}
Across the eight configurations, the mean Avg.\ Dice on slice~577 differs from that on slice~627 by $+0.0001$ raw, $+0.0012$ post-processed and $+0.0003$ with TTA3. Single configurations change by \dn{0.0016} to $+0.0032$ raw, \dn{0.0010} to $+0.0040$ post-processed and \dn{0.0012} to $+0.0033$ with TTA3, so no configuration moves by more than 0.0040. All changes lie below the 0.005 we read as real (Section~\mainref{sec:seeds}). The final configuration scores 0.8910 raw, 0.8980 with TTA3 and 0.9080 post-processed on slice~577, which is $+0.0007$, $+0.0003$ and $+0.0007$ from its values on slice~627. The ranking of Table~\mainref{tab:config_selection} does not carry over exactly: on slice~577 the final configuration is second raw and second post-processed, behind \muvitenc{} with $65{\times}65$ views and a $5{\times}5$ kernel by 0.0006 and 0.0011, and first with TTA3, by 0.0001. All three margins are below 0.005, so the two configurations are not separated on either slice.

\paragraph{TTA3 and Post-Processing.}
Both inference steps help on slice~577 as on slice~627. For the final configuration, post-processing raises the raw Avg.\ Dice by 0.0170 on both slices. TTA3 raises the mean raw Avg.\ Dice of the eight configurations by 0.0071 (0.0069 on slice~627), and post-processing by 0.0206 (0.0194). The post-processing was tuned on slice~627 and not on slice~577, so its gain on slice~577 is not inflated by the tuning: the model performs as well on a slice that played no part in the tuning.

\paragraph{Scope.}
Slice~577 comes from the same cell as slice~627, so the two slices share the cell's appearance, its labels and its mask. One further slice is not a mean over slices, and it says nothing about other cells. The seeds do not resample the sparse label set, so the SDs exclude that variance, as on slice~627.

\subsection{Slices Across the Nucleus Range}
\label{sec:appendix-volume}

Slices 577 and 627 are two slices of one cell. To see how the score varies along the cell, we scored one trained model, seed~74 of the final configuration, on further slices of high\_c4 with the same checkpoint, TTA3, post-processing and metric as before (Section~\mainref{sec:configs}); nothing was retrained or retuned. We took every 25th slice from 251 to 651, the part of the range in which the cell contains the nucleus class. This gives 17 slices, 9 of them in steps of 50 and 8 in between. Slices 577 and 627 are not among them (they are scored in \cref{tab:slice577} and in the main paper); rows 576 and 626 lie next to them and are nearly the same slices. Avg.\ Dice is the mean of the four per-class Dice scores, so it is defined only for slices that contain all four classes. Slices outside this range lack the nucleus class, and the score of a missing class (zero if a few pixels are predicted, undefined otherwise) would dominate the mean. \Cref{tab:volume} gives the results.

\input{fig/tab_volume}

\paragraph{Scores.}
Over the 17 slices the mean Avg.\ Dice is $0.8838\pm0.0152$ raw, $0.8908\pm0.0145$ with TTA3 and $0.9038\pm0.0114$ with TTA3 and post-processing (mean $\pm$ SD over slices), 0.010 and 0.009 below seed~74 on slice~627 (0.9005 and 0.9124 in \cref{tab:perclass}). Nucleus is the best class (0.9747 with TTA3), granule the lowest (0.8043), as on slice~627. Mitochondria varies the most between slices (SD 0.0378 with TTA3, range 0.7895 to 0.9175). The lowest Avg.\ Dice with TTA3 is on slice~351 (0.8600) and the highest on slice~476 (0.9126). Scores are lower at the ends of the range, where less of the slice is labeled: the first slices have 165,588 and 197,603 in-cell pixels, against about 310,000 in the middle.

\paragraph{Scope.}
The slices come from the one cell on which the configuration was selected, so they say how the score varies along that cell and not how it transfers to other cells. Neighboring slices are correlated, so the SD describes variation along the cell and 17 slices are not 17 independent samples. The slices are chosen from the ground truth (those that contain the nucleus), and one seed was scored, so the SD excludes the variance between seeds.

\section{Limitations in Detail}
\label{sec:appendix-limitations}

The Discussion and Conclusion of the main paper (Section~\mainref{sec:limitations}) states the main limitations briefly. This part gives the details.

\paragraph{Evaluation.}
Our results rest on one slice of one cell, which cuts through a single nucleus; a second slice of the same cell is scored in \cref{sec:appendix-slice577}. Only high-glucose $\beta$ cells of one dataset are tested. We made development choices on this slice, and the post-processing was tuned on it (ranked on its left half with earlier models, without TTA3). The scores of the final configuration are optimistic because it is the best of eight (\cref{sec:appendix-optimism}). Table~\mainref{tab:comparison} shows the three-seed mean of \epssegformer{} against single baseline values. The ablations are single runs, and the seeds do not resample the sparse labels, so the SDs exclude that variance. Avg.\ Dice is a pixel measure and ignores Granule instances.

\paragraph{Method.}
Keeping the center token visible in every view lowers Avg.\ Dice in all three seeds, by 0.0077 on average (\cref{tab:center_full}); the other evidence is the single-view comparison of \cref{tab:masking-summary} and the argument of Section~\mainref{sec:hidden} of the main paper. We have not isolated the effect of the center token being visible at inference (Section~\mainref{sec:inference}); training never shows it. Coarse views are sampled without smoothing, so aliasing is possible. The coarsest view always extends beyond the evaluation slice (Section~\mainref{sec:res-augpad}), so the padded area could let the model infer its position relative to the slice border; we have not tested this. The model ignores context along $z$, although the voxels are isotropic.

\paragraph{Use on New Data.}
Our labels are exact and class-balanced (Section~\mainref{sec:data}). The pipeline needs the cell mask: in-cell pixels come from the ground truth, the model has no outside-cell class, and post-processing uses the mask too. A new volume would first need a segmentation of each cell, and predicted masks are untested. Inference is slow: with one forward pass per pixel, the present code takes about 1.6\,GPU-hours on a V100 for slice~627 and about 41\,GPU-days for the in-cell voxels of a whole cell (extrapolated by pixel count), and TTA3 triples both (Part~\ref{sec:appendix-setup}, Inference Cost). The code also runs the decoders (about 13\% of the per-pixel FLOPs), although the prediction does not need them.

%% file: fig/tab_muvit_diff.tex
\begin{tablewide}[p]
\centering
\caption{\textbf{The \epssegformer{} Architecture Compared With MuViT.} The top block lists the design that the two models share. The blocks below list the differences by component between the final model (Ours) and the official MuViT implementation; code pointers (file:line) refer to the repository of the latter.}
\label{tab:muvit_diff}
\tabstyledbl
\newcolumntype{L}[1]{>{\raggedright\arraybackslash}p{#1}}
\newcommand{\cp}[1]{{\texttt{\scriptsize #1}}}
\newlength{\wlab}\newlength{\wval}
\setlength{\wlab}{0.15\textwidth}
\setlength{\wval}{\dimexpr(\textwidth-\wlab-6\tabcolsep)/2\relax} %
\begin{tabular}{L{\wlab} L{\wval} L{\wval}}
\toprule
\grouprow{3}{Shared with MuViT}
Blocks & \multicolumn{2}{L{\dimexpr2\wval+2\tabcolsep\relax}}{Pre-LN transformer blocks with full attention across all resolution levels.} \\
Positions & \multicolumn{2}{L{\dimexpr2\wval+2\tabcolsep\relax}}{2D RoPE on world coordinates.} \\
Masking & \multicolumn{2}{L{\dimexpr2\wval+2\tabcolsep\relax}}{Masked tokens are dropped from the encoder and re-inserted before the decoder.} \\
Reconstruction & \multicolumn{2}{L{\dimexpr2\wval+2\tabcolsep\relax}}{Masked-patch reconstruction loss on raw pixels.} \\
Per-level decoder & \multicolumn{2}{L{\dimexpr2\wval+2\tabcolsep\relax}}{Linear--GELU--LayerNorm input projection and two layers: layer 1 cross-attends to all levels, layer 2 self-attends within the level; a linear unpatchify layer ends the decoder.} \\
\midrule
 & \textbf{Ours} & \textbf{MuViT} \\
\cmidrule(lr){2-2}\cmidrule(lr){3-3}
\grouprow{3}{Encoder}
Size & 14 layers, width 192, 1 head, MLP ratio 2 & Defaults: width 320, 4 heads (\cp{encoders.py:22}), feed-forward multiplier 4 (\cp{bblocks.py:173}) \\
Patch embedding & One shared Conv2d with bias, no norm & Per-level Linear + LayerNorm \\
Level embedding & \texttt{nn.Embedding}, truncated normal (std 0.02), added to the raw patch embedding & \texttt{randn} parameter (std 1), added after the LayerNorm \\
Final encoder norm & LayerNorm on the encoder output & None \\
Attention projections & With bias & Bias-free (\texttt{x\_transformers} Attention) \\
Token grid & Cell centers, $(i+0.5)/g$ & \texttt{linspace(0,1,h)}, endpoints included \\
RoPE frequencies & Fixed, base $10^4$ & Trainable \texttt{inv\_freq}, default mode \texttt{per\_layer} \\
RoPE coordinates & World coordinates divided by the view width $W$ (85\,px in the final model) &Normalized $[0,1]$ coordinates \\
\midrule
\grouprow{3}{Masking}
Ratio & Stepped schedule 0.70, 0.55, 0.40, then 0.25, changing every 5 epochs & Constant 0.75 \\
Level allocation & Uniform, no Dirichlet & Dirichlet(0.5) level weights \\
Mask token & One shared token (truncated normal, std 0.02) & \texttt{randn}, optionally per level \\
Forced masking & The center token of every level is always masked for labeled samples & None \\
\midrule
\grouprow{3}{Decoder and loss}
Decoder size & Width 192, 1 head & Default width 256, 8 heads (\cp{mae.py:43}) \\
Final projection & Separate linear layer per level & One shared linear layer (\cp{mae.py:125}) \\
Decoder RoPE and projections & Fixed RoPE, projections with bias & Trainable RoPE, bias-free projections \\
Loss & MSE per level, then a plain mean over levels & MSE pooled over all masked tokens (\cp{mae.py:311}) \\
Extra loss terms & No FFT term, no patch normalization & Optional \texttt{norm\_mse} and \texttt{mse\_fft} modes; the default is plain MSE \\
\midrule
\grouprow{3}{Training and task}
Optimizer & AdamW, lr $10^{-4}$, weight decay $10^{-4}$, no schedule & AdamW, betas $(0.9, 0.95)$, lr $=10^{-4}\sqrt{\text{bs}\cdot\text{gpus}/128}$ \\
Schedule, precision & None & 2 warmup epochs, gradient clipping at 0.5, bf16 mixed precision \\
Task & Joint cross-entropy and reconstruction, with the context-only head and a $5{\times}5$ kernel & Masked pre-training only \\
Test time & Nothing masked; TTA3 (slice and two flips); borders padded with the mean intensity & Not part of the pipeline \\
\bottomrule
\end{tabular}
\end{tablewide}

%% file: fig/tab_data.tex
\begin{tablewide}[!t]
\centering
\caption{\textbf{BetaSeg Cells: c1--c3 for Training, high\_c4 for Evaluation.} Shape is the size of the volume in voxels ($z\times y\times x$); we process 2D $y\times x$ slices. In-cell is the share of voxels inside the segmented cell. Class frequency is the share of in-cell voxels of each class, computed over the whole volume except in the last row. Slices counts the training and validation slices that contain the cell (the validation slices are a 15\% hold-out for checkpoint selection). Labels are the simulated pixel labels; validation labels are drawn in the same way at about twice the density. The c1--c3 row pools the three cells. Slice 627 is index 626 when slices are counted from 0. Unc., Nuc., Gran. and Mito. stand for Uncategorized, Nucleus, Granule and Mitochondria.}
\label{tab:data}
\tabstyledbl
\begin{tabular*}{\textwidth}{@{\extracolsep{\fill}}llcrrrrcrr}
\toprule
 & & & \multicolumn{4}{c}{Class frequency (\%)} & & \multicolumn{2}{c}{Labels} \\
\cmidrule(lr){4-7}\cmidrule(lr){9-10}
Cell & Shape ($z{\times} y{\times} x$) & In-cell (\%) & \multicolumn{1}{r}{Unc.} & \multicolumn{1}{r}{Nuc.} & \multicolumn{1}{r}{Gran.} & \multicolumn{1}{r}{Mito.} & Slices (train/val) & \multicolumn{1}{r}{Training} & \multicolumn{1}{r}{Validation} \\
\midrule
c1 & $1097{\times}699{\times}760$ & 42.3 & 60.3 & 12.5 & 20.5 & 6.7 & 912/162 & 3,399 & 1,236 \\
c2 & $1043{\times}606{\times}870$ & 37.0 & 58.0 & 13.9 & 18.4 & 9.7 & 841/149 & 3,302 & 1,178 \\
c3 & $1023{\times}676{\times}845$ & 37.5 & 55.4 & 11.4 & 28.2 & 4.9 & 858/152 & 3,328 & 1,182 \\
\midrule
c1--c3 & & 39.0 & 58.0 & 12.6 & 22.4 & 7.0 & 2,611/463 & 10,029 & 3,596 \\
\midrule
high\_c4 (evaluation) & $1022{\times}545{\times}1082$ & 32.1 & 55.6 & 13.4 & 19.1 & 11.9 & & none & none \\
high\_c4, slice 627 & $1{\times}545{\times}1082$ & 53.8 & 57.2 & 12.8 & 17.9 & 12.1 & & none & none \\
\bottomrule
\end{tabular*}
\end{tablewide}

%% file: fig/tab_aug_params.tex
\begin{table}[t]
\centering
\caption{\textbf{Augmentation Used in All Augmented Runs.} A sample is augmented with probability $0.8$, and each row of the table is then applied independently with its own probability. Geometric transformations move the sampling positions of all views around the center pixel (bilinear interpolation for the image, nearest neighbor for the label patch); positions outside the slice keep the constant padding value. Intensity changes act on the z-scored data in the order gamma, contrast and brightness, blur, noise. Blur and noise are given for the finest view and divided by the stride $s_\ell$ of view $\ell$. The view width is $W=85$ slice pixels, and $U(a,b)$ denotes the uniform distribution on $[a,b]$.}
\label{tab:aug_params}
\tabstylenarrow
\begin{tabular*}{\columnwidth}{@{\extracolsep{\fill}}lc>{\raggedright\arraybackslash}p{0.50\columnwidth}}
\toprule
Transformation & Probability & Parameters \\
\midrule
Any augmentation & 0.8 & -- \\
\midrule
\grouprow{3}{Geometry}
Flip (one axis) & 0.5 & -- \\
Quarter turns & 0.5 & $k\in\{1,2,3\}$ turns of $90^\circ$, uniform \\
Rotation & 0.3 & angle $U(-15^\circ,15^\circ)$ \\
Elastic deformation & 0.3 & $4{\times}4$ control grid, displacement $\mathcal N(0,\sigma^2)$ per component, $\sigma=0.03\,W=2.55$ (slice px), zero at the center \\
\midrule
\grouprow{3}{Intensity}
Brightness & \multirow{3}{*}{0.8} & shift $U(-0.4,0.4)$ (z-scored units) \\
Contrast &  & factor $U(0.6,1.4)$ \\
Gamma &  & signed power $U(0.8,1.25)$ \\
Gaussian blur & 0.2 & $\sigma\sim U(0.1,2.0)$ (slice px) \\
Gaussian noise & 0.3 & $\sigma\sim U(0.01,0.1)$ (z-scored units) \\
\bottomrule
\end{tabular*}
\end{table}

%% file: fig/tab_perclass.tex
\begin{table}[!t]
\centering
\caption{\textbf{Granule Is the Lowest Class in \mbox{Every} Seed, With and Without Post-Processing.} Per-class Dice of the final model with TTA3, for each of the three seeds. The Mean rows give the average over the seeds, with the sample standard deviation (SD) of Avg.\ Dice below it. Slice 627 of cell high\_c4 has 317,438 in-cell pixels: Nuc.\ 40,555, Gran.\ 56,758, Mito.\ 38,443 and Unc.\ 181,682.}
\label{tab:perclass}
\tabstylenarrow
\newcommand{\classtab}{%
\begin{tabular}{lccccc}%
\toprule%
Seed & Nuc. & Gran. & Mito. & Unc. & Avg.\ Dice \\%
\midrule%
\grouprow{6}{TTA3 and post-processing}%
54  & 0.9794 & 0.8169 & 0.9201 & 0.9279 & 0.9111 \\%
74  & 0.9782 & 0.8169 & 0.9274 & 0.9270 & 0.9124 \\%
313 & 0.9757 & 0.8023 & 0.9173 & 0.9186 & 0.9035 \\%
\midrule%
\emphrow Mean & 0.9778 & 0.8120 & 0.9216 & 0.9245 & \textbf{0.9090} \\[-1.5pt]%
\emphrow & & & & & \sd{0.0048} \\%
\midrule%
\grouprow{6}{TTA3, no post-processing}%
54  & 0.9765 & 0.7994 & 0.9127 & 0.9166 & 0.9013 \\%
74  & 0.9753 & 0.7997 & 0.9113 & 0.9157 & 0.9005 \\%
313 & 0.9735 & 0.7805 & 0.9064 & 0.9045 & 0.8912 \\%
\midrule%
Mean & 0.9751 & 0.7932 & 0.9101 & 0.9122 & 0.8977 \\[-1.5pt]%
 & & & & & \sd{0.0056} \\%
\bottomrule%
\end{tabular}}
\fitwidth{\columnwidth}{12}{\classtab}
\end{table}

%% file: fig/tab_center_full.tex
\begin{table}[!t]
\centering
\caption{\textbf{Keeping the Center Token Visible in Every View Lowers Avg.\ Dice in All Three Seeds.} Avg.\ Dice of a single forward pass (Base), MuViT with $85{\times}85$ views, $5{\times}5$ patches and a $5{\times}5$ S.H.\ kernel, summarized in Section~\mainref{sec:res-scales} of the main paper. Base is the main configuration proposed in the paper (the $85{\times}85$, $5{\times}5$ row of \cref{tab:config_full}), which drops the center token of every view; in the visible runs the center tokens stay visible in training, so the S.H.\ reads one more token. $\Delta$ is visible minus Base. Mean $\pm$ sample SD ($n-1$).}
\label{tab:center_full}
\tabstylenarrow
\begin{tabular*}{\columnwidth}{@{\extracolsep{\fill}}lcccc}
\toprule
\makecell[l]{Center token\\in every view} & 54 & 74 & 313 & Mean\,$\pm$\,SD \\
\midrule
Base & 0.8951 & 0.8938 & 0.8819 & \msd{0.8903}{0.0073} \\
Visible & 0.8770 & 0.8903 & 0.8805 & \msd{0.8826}{0.0069} \\
\midrule
$\Delta$ & \dn{0.0181} & \dn{0.0035} & \dn{0.0014} & \dn{0.0077} \\
\bottomrule
\end{tabular*}
\end{table}

%% file: fig/tab_aug_ctx.tex
\begin{table}[!t]
\centering
\caption{\textbf{Augmentation Raises Avg.\ Dice in All Eight Settings.} Both architectures are trained without border padding: \muvitenc{} has one CNN decoder and MuViT has one transformer decoder per view. $\Delta$ is the Avg.\ Dice with augmentation minus that without it, and Mean is the average over the four settings of each architecture. Each value is a single run.}
\label{tab:aug_ctx}
\tabstylenarrow
\begin{tabular*}{\columnwidth}{@{\extracolsep{\fill}}llccc}
\toprule
 & & \multicolumn{2}{c}{Avg.\ Dice} & \\
\cmidrule(lr){3-4}
\makecell[l]{View / patch\\size (px)} & \makecell[c]{S.H.\\kernel} & Without aug. & With aug. & $\Delta$ \\
\midrule
\grouprow{5}{\muvitenc}
\multirow{2}{*}{$27{\times}27$ / $3{\times}3$} & $3{\times}3$ & 0.7999 & 0.8762 & $+0.0763$ \\
 & $5{\times}5$ & 0.8194 & 0.8738 & $+0.0544$ \\
\cmidrule(lr){1-5}
\multirow{2}{*}{$65{\times}65$ / $5{\times}5$} & $3{\times}3$ & 0.7593 & 0.8507 & $+0.0914$ \\
 & $5{\times}5$ & 0.7829 & 0.8630 & $+0.0801$ \\
\midrule
\multicolumn{2}{l}{Mean} & 0.7904 & 0.8659 & $+0.0755$ \\
\midrule
\grouprow{5}{MuViT}
\multirow{2}{*}{$27{\times}27$ / $3{\times}3$} & $3{\times}3$ & 0.8298 & 0.8771 & $+0.0473$ \\
 & $5{\times}5$ & 0.8130 & 0.8837 & $+0.0707$ \\
\cmidrule(lr){1-5}
\multirow{2}{*}{$65{\times}65$ / $5{\times}5$} & $3{\times}3$ & 0.7631 & 0.8712 & $+0.1081$ \\
 & $5{\times}5$ & 0.7977 & 0.8790 & $+0.0813$ \\
\midrule
\multicolumn{2}{l}{Mean} & 0.8009 & 0.8778 & $+0.0769$ \\
\bottomrule
\end{tabular*}
\end{table}

%% file: fig/tab_padding.tex
\begin{table}[!t]
\centering
\caption{\textbf{Border Padding Raises Avg.\ Dice for \muvitenc{} in All Four Settings; for MuViT the Changes Are Below 0.005.} All runs use augmentation. Without padding, pixels whose views extend beyond the image are excluded from training. With padding, the views are completed with padded pixels, so these pixels are kept. $\Delta$ is the Avg.\ Dice with padding minus that without it, and Mean is the average over the four settings. Each value is a single run.}
\label{tab:padding}
\tabstylenarrow
\begin{tabular*}{\columnwidth}{@{\extracolsep{\fill}}llccc}
\toprule
 & & \multicolumn{2}{c}{Avg.\ Dice} & \\
\cmidrule(lr){3-4}
\makecell[l]{View / patch\\size (px)} & \makecell[c]{S.H.\\kernel} & Without pad. & With pad. & $\Delta$ \\
\midrule
\grouprow{5}{\muvitenc}
\multirow{2}{*}{$27{\times}27$ / $3{\times}3$} & $3{\times}3$ & 0.8762 & 0.8843 & $+0.0081$ \\
 & $5{\times}5$ & 0.8738 & 0.8811 & $+0.0073$ \\
\cmidrule(lr){1-5}
\multirow{2}{*}{$65{\times}65$ / $5{\times}5$} & $3{\times}3$ & 0.8507 & 0.8768 & $+0.0261$ \\
 & $5{\times}5$ & 0.8630 & 0.8873 & $+0.0243$ \\
\midrule
\multicolumn{2}{l}{Mean} & 0.8659 & 0.8824 & $+0.0165$ \\
\midrule
\grouprow{5}{MuViT}
\multirow{2}{*}{$27{\times}27$ / $3{\times}3$} & $3{\times}3$ & 0.8771 & 0.8742 & \dn{0.0029} \\
 & $5{\times}5$ & 0.8837 & 0.8842 & $+0.0005$ \\
\cmidrule(lr){1-5}
\multirow{2}{*}{$65{\times}65$ / $5{\times}5$} & $3{\times}3$ & 0.8712 & 0.8743 & $+0.0031$ \\
 & $5{\times}5$ & 0.8790 & 0.8822 & $+0.0032$ \\
\midrule
\multicolumn{2}{l}{Mean} & 0.8778 & 0.8787 & $+0.0009$ \\
\bottomrule
\end{tabular*}
\end{table}

%% file: fig/tab_padding_pixels.tex
\begin{table}[!t]
\centering
\caption{\textbf{Border Padding Adds Training Labels and Changes the Scored Pixels by Less Than 2\%; on Identical Pixels the Gain Is Slightly Larger.} These are the 16 runs of \cref{tab:padding} (scales $\{1,2,4\}$, augmentation). \textbf{(a)}~Without padding, training and validation labels whose coarsest view leaves the slice are not used, and scoring omits the in-cell pixels within half the finest view of the slice border (13\,px at $27{\times}27$, 32\,px at $65{\times}65$). \textbf{(b)}~$\Delta$ is the Avg.\ Dice with padding minus that without it: Published is that of \cref{tab:padding}, and Same pixels scores both models on the pixels the unpadded model scores. Mean averages the four settings of one architecture. Each value is a single run.}
\label{tab:padding_pixels}
\tabstylenarrow
\begin{tabular*}{\columnwidth}{@{\extracolsep{\fill}}lcc}
\toprule
\panelrow{3}{(a) Labels and scored pixels}
 & $27{\times}27$ & $65{\times}65$ \\
\cmidrule(lr){2-3}
Training labels, without pad. & 9,922 & 8,365 \\
Training labels, with pad. & 10,029 & 10,029 \\
\hspace{1em}Added by padding & $+107$ ($+1.1\%$) & $+1{,}664$ ($+19.9\%$) \\
Validation labels, without pad. & 3,551 & 3,022 \\
Validation labels, with pad. & 3,596 & 3,596 \\
In-cell pixels scored, without pad. & 315,469 & 311,821 \\
In-cell pixels scored, with pad. & 317,438 & 317,438 \\
\hspace{1em}Not scored without padding & 1,969 ($0.62\%$) & 5,617 ($1.77\%$) \\
\midrule
\panelrow{3}{(b) $\Delta$ Avg.\ Dice from padding}
\makecell[l]{View size,\\S.H.\ kernel} & Published & Same pixels \\
\cmidrule(lr){1-3}
\grouprow{3}{\muvitenc}
$27{\times}27$, $3{\times}3$ & $+0.0081$ & $+0.0110$ \\
$27{\times}27$, $5{\times}5$ & $+0.0073$ & $+0.0092$ \\
$65{\times}65$, $3{\times}3$ & $+0.0261$ & $+0.0287$ \\
$65{\times}65$, $5{\times}5$ & $+0.0243$ & $+0.0263$ \\
Mean & $+0.0165$ & $+0.0188$ \\
\midrule
\grouprow{3}{MuViT}
$27{\times}27$, $3{\times}3$ & \dn{0.0029} & \dn{0.0011} \\
$27{\times}27$, $5{\times}5$ & $+0.0005$ & $+0.0022$ \\
$65{\times}65$, $3{\times}3$ & $+0.0031$ & $+0.0042$ \\
$65{\times}65$, $5{\times}5$ & $+0.0032$ & $+0.0047$ \\
Mean & $+0.0010$ & $+0.0025$ \\
\bottomrule
\end{tabular*}
\end{table}

%% file: fig/tab_scales1_full.tex
\begin{table}[!t]
\centering
\caption{\textbf{Removing the Two Coarser Views Lowers Avg.\ Dice in All 18 Seed Pairs.} Avg.\ Dice of a single forward pass for MuViT and \muvitenc{} with a $5{\times}5$ S.H.\ kernel, trained with the finest view only (scale $\{1\}$) or with the three views of the main configuration (scales $\{1,4,8\}$, the rows with a $5{\times}5$ kernel of \cref{tab:config_full}), summarized in Section~\mainref{sec:res-scales} of the main paper. All other settings are the same, and the single-view runs have a third of the tokens. $\Delta$ is the scales $\{1,4,8\}$ minus $\{1\}$. Mean $\pm$ sample SD ($n-1$) over seeds 54, 74 and 313.}
\label{tab:scales1_full}
\tabstylenarrow
\begin{tabular*}{\columnwidth}{@{\extracolsep{\fill}}llcccc}
\toprule
\makecell[c]{View /\\patch (px)} & Scales & 54 & 74 & 313 & Mean\,$\pm$\,SD \\
\midrule
\grouprow{6}{MuViT}
\multirow{3}{*}{\makecell[c]{$27{\times}27$\\$3{\times}3$}} & $\{1\}$ & 0.8375 & 0.8531 & 0.8540 & \msd{0.8482}{0.0093} \\
 & $\{1,4,8\}$ & 0.8702 & 0.8778 & 0.8662 & \msd{0.8714}{0.0059} \\
 & $\Delta$ & $+0.0327$ & $+0.0247$ & $+0.0123$ & $+0.0232$ \\
\cmidrule(lr){1-6}
\multirow{3}{*}{\makecell[c]{$65{\times}65$\\$5{\times}5$}} & $\{1\}$ & 0.8627 & 0.8635 & 0.8621 & \msd{0.8627}{0.0007} \\
 & $\{1,4,8\}$ & 0.8882 & 0.8843 & 0.8786 & \msd{0.8837}{0.0048} \\
 & $\Delta$ & $+0.0255$ & $+0.0208$ & $+0.0165$ & $+0.0210$ \\
\cmidrule(lr){1-6}
\multirow{3}{*}{\makecell[c]{$85{\times}85$\\$5{\times}5$}} & $\{1\}$ & 0.8641 & 0.8605 & 0.8392 & \msd{0.8546}{0.0134} \\
 & $\{1,4,8\}$ & 0.8951 & 0.8938 & 0.8819 & \msd{0.8903}{0.0073} \\
 & $\Delta$ & $+0.0310$ & $+0.0333$ & $+0.0427$ & $+0.0357$ \\
\midrule
\grouprow{6}{\muvitenc}
\multirow{3}{*}{\makecell[c]{$27{\times}27$\\$3{\times}3$}} & $\{1\}$ & 0.8478 & 0.8537 & 0.8529 & \msd{0.8514}{0.0032} \\
 & $\{1,4,8\}$ & 0.8891 & 0.8748 & 0.8885 & \msd{0.8841}{0.0081} \\
 & $\Delta$ & $+0.0414$ & $+0.0212$ & $+0.0355$ & $+0.0327$ \\
\cmidrule(lr){1-6}
\multirow{3}{*}{\makecell[c]{$65{\times}65$\\$5{\times}5$}} & $\{1\}$ & 0.8747 & 0.8726 & 0.8710 & \msd{0.8728}{0.0019} \\
 & $\{1,4,8\}$ & 0.8874 & 0.8911 & 0.8866 & \msd{0.8884}{0.0024} \\
 & $\Delta$ & $+0.0127$ & $+0.0185$ & $+0.0156$ & $+0.0156$ \\
\cmidrule(lr){1-6}
\multirow{3}{*}{\makecell[c]{$85{\times}85$\\$5{\times}5$}} & $\{1\}$ & 0.8700 & 0.8654 & 0.8591 & \msd{0.8649}{0.0055} \\
 & $\{1,4,8\}$ & 0.8853 & 0.8870 & 0.8852 & \msd{0.8858}{0.0010} \\
 & $\Delta$ & $+0.0153$ & $+0.0216$ & $+0.0261$ & $+0.0210$ \\
\bottomrule
\end{tabular*}
\end{table}

%% file: fig/tab_scales_full.tex
\begin{table}[!t]
\centering
\caption{\textbf{Crop Scales $\{1,2,4\}$ Versus $\{1,4,8\}$ for Every S.H.\ Kernel Size.} Avg.\ Dice with augmentation and border padding. $\Delta$ is $\{1,4,8\}$ minus $\{1,2,4\}$, and (all) marks a kernel that covers the whole patch grid. Bold marks the best S.H.\ kernel per scale set within each view / patch size and architecture. Each value is a single run.}
\label{tab:scales_full}
\tabstylenarrow
\begin{tabular*}{\columnwidth}{@{\extracolsep{\fill}}llccc}
\toprule
 & & \multicolumn{2}{c}{Avg.\ Dice, scales} & \\
\cmidrule(lr){3-4}
\makecell[l]{View / patch\\size (px)} & \makecell[c]{S.H.\\kernel} & $\{1,2,4\}$ & $\{1,4,8\}$ & $\Delta$ \\
\midrule
\grouprow{5}{\muvitenc}
\multirow{3}{*}{$27{\times}27$ / $3{\times}3$} & $3{\times}3$ & \textbf{0.8843} & \textbf{0.8865} & $+0.0022$ \\
 & $5{\times}5$ & 0.8811 & 0.8843 & $+0.0032$ \\
 & $9{\times}9$ (all) & 0.8389 & 0.8523 & $+0.0134$ \\
\cmidrule(lr){1-5}
\multirow{4}{*}{$65{\times}65$ / $5{\times}5$} & $3{\times}3$ & 0.8768 & 0.8712 & \dn{0.0056} \\
 & $5{\times}5$ & \textbf{0.8873} & \textbf{0.8889} & $+0.0016$ \\
 & $9{\times}9$ & 0.8631 & 0.8628 & \dn{0.0003} \\
 & $13{\times}13$ (all) & 0.8314 & 0.8242 & \dn{0.0072} \\
\midrule
\grouprow{5}{MuViT}
\multirow{3}{*}{$27{\times}27$ / $3{\times}3$} & $3{\times}3$ & 0.8742 & 0.8852 & $+0.0110$ \\
 & $5{\times}5$ & \textbf{0.8842} & \textbf{0.8894} & $+0.0052$ \\
 & $9{\times}9$ (all) & 0.8356 & 0.8300 & \dn{0.0056} \\
\cmidrule(lr){1-5}
\multirow{4}{*}{$65{\times}65$ / $5{\times}5$} & $3{\times}3$ & 0.8743 & 0.8760 & $+0.0017$ \\
 & $5{\times}5$ & \textbf{0.8822} & \textbf{0.8856} & $+0.0034$ \\
 & $9{\times}9$ & 0.8594 & 0.8651 & $+0.0057$ \\
 & $13{\times}13$ (all) & 0.8087 & 0.8303 & $+0.0216$ \\
\bottomrule
\end{tabular*}
\end{table}

%% file: fig/tab_viewsize_full.tex
\begin{tablewide}[!t]
\centering
\caption{\textbf{Smaller Views Win Before Augmentation; After It, the Best View Size Depends on Architecture and S.H.\ Kernel.} Panel (a) uses scales $\{1,2,4\}$ and no border padding; panel (b) uses scales $\{1,4,8\}$ and border padding. The patch grid is the view size divided by the patch size, and Tokens is the number of input tokens of the three views ($3\times$ the patch grid area). $\Delta$ is the change from the next smaller view in the same panel, for the same model and kernel. Bold marks the best view size per panel and model. Each value is a single run.}
\label{tab:viewsize_full}
\tabstylesparse
\begin{tabular*}{\textwidth}{@{\extracolsep{\fill}}lccccccc}
\toprule
 & & & & \multicolumn{2}{c}{\muvitenc} & \multicolumn{2}{c}{MuViT} \\
\cmidrule(lr){5-6}\cmidrule(lr){7-8}
\makecell[l]{View / patch\\size (px)} & \makecell[c]{Patch\\grid} & Tokens & \makecell[c]{S.H.\\kernel} & Avg.\ Dice & $\Delta$ & Avg.\ Dice & $\Delta$ \\
\midrule
\panelrow{8}{(a) Before augmentation, $3{\times}3$ patches}
\multirow{2}{*}{$27{\times}27$ / $3{\times}3$} & \multirow{2}{*}{$9{\times}9$} & \multirow{2}{*}{243} & $3{\times}3$ & 0.7999 & -- & \textbf{0.8298} & -- \\
 & & & $5{\times}5$ & \textbf{0.8194} & -- & 0.8130 & -- \\
\cmidrule(lr){1-8}
\multirow{2}{*}{$63{\times}63$ / $3{\times}3$} & \multirow{2}{*}{$21{\times}21$} & \multirow{2}{*}{1323} & $3{\times}3$ & 0.7719 & \dn{0.0280} & 0.7787 & \dn{0.0511} \\
 & & & $5{\times}5$ & 0.8069 & \dn{0.0125} & 0.7933 & \dn{0.0197} \\
\midrule
\panelrow{8}{(b) After augmentation, $5{\times}5$ patches}
\multirow{2}{*}{$65{\times}65$ / $5{\times}5$} & \multirow{2}{*}{$13{\times}13$} & \multirow{2}{*}{507} & $3{\times}3$ & 0.8712 & -- & 0.8760 & -- \\
 & & & $5{\times}5$ & 0.8889 & -- & 0.8856 & -- \\
\cmidrule(lr){1-8}
\multirow{2}{*}{$85{\times}85$ / $5{\times}5$} & \multirow{2}{*}{$17{\times}17$} & \multirow{2}{*}{867} & $3{\times}3$ & \textbf{0.8911} & $+0.0199$ & 0.8846 & $+0.0086$ \\
 & & & $5{\times}5$ & 0.8819 & \dn{0.0070} & \textbf{0.8882} & $+0.0026$ \\
\cmidrule(lr){1-8}
\multirow{2}{*}{$105{\times}105$ / $5{\times}5$} & \multirow{2}{*}{$21{\times}21$} & \multirow{2}{*}{1323} & $3{\times}3$ & 0.8774 & \dn{0.0137} & 0.8793 & \dn{0.0053} \\
 & & & $5{\times}5$ & 0.8887 & $+0.0068$ & 0.8811 & \dn{0.0071} \\
\bottomrule
\end{tabular*}
\end{tablewide}

%% file: fig/tab_config_full.tex
\begin{tablewide}[!t]
\centering
\caption{\textbf{Per-Seed Results of the Eight Configurations of Table~\mainref{tab:config_selection}, With Epochs and Parameter Counts.} Avg.\ Dice of the checkpoint with the lowest validation loss, for seeds 54, 74 and 313; Epoch is the epoch of that checkpoint. SD is the sample standard deviation ($n-1$). Raw is a single forward pass without post-processing, and Post is post-processing without TTA3. For TTA3 and TTA3 + Post only the mean and SD exist (Section~\mainref{sec:inference}). Params is the number of trained parameters, including the decoder(s). Bold marks the best mean of the eight configurations per measure, underline the best \muvitenc{}, and shading the final configuration. $^\dagger$: $27{\times}27$ views from an earlier set, not ranked, without TTA3.}
\label{tab:config_full}
\tabstyledbl\renewcommand{\arraystretch}{1.2} %
\newcommand{\cfgtab}{%
\begin{tabular}{lrcr >{\hspace{6pt}}cccc >{\hspace{2pt}}c >{\hspace{6pt}}cccc >{\hspace{2pt}}c >{\hspace{6pt}}rrrr}%
\toprule%
 & & & & \multicolumn{4}{c}{Raw} & TTA3 & \multicolumn{4}{c}{Post} & \makecell{TTA3\\[-1pt]+\,Post} & \multicolumn{4}{c}{Epoch} \\%
\cmidrule(lr){5-8}\cmidrule(lr){9-9}\cmidrule(lr){10-13}\cmidrule(lr){14-14}\cmidrule(lr){15-18}%
\makecell[l]{View / patch\\[-1pt]size (px)} & Tokens & \makecell{S.H.\\[-1pt]kernel} & \makecell{Params\\[-1pt](M)} & 54 & 74 & 313 & \makecell{Mean\\[-1pt]$\pm$ SD} & \makecell{Mean\\[-1pt]$\pm$ SD} & 54 & 74 & 313 & \makecell{Mean\\[-1pt]$\pm$ SD} & \makecell{Mean\\[-1pt]$\pm$ SD} & 54 & 74 & 313 & Mean \\%
\midrule%
\grouprow{18}{MuViT}%
$65{\times}65$ / $5{\times}5$ & 507 & $3{\times}3$ & 6.47 & 0.8911 & 0.8791 & 0.8756 & 0.8819 & 0.8889 & 0.9116 & 0.9055 & 0.8996 & 0.9056 & 0.9076 & 210 & 166 & 196 & 190.7 \\[-1.5pt]%
 &  &  &  &  &  &  & \sd{0.0081} & \sd{0.0094} &  &  &  & \sd{0.0060} & \sd{0.0069} &  &  &  &  \\%
 &  & $5{\times}5$ & 7.25 & 0.8882 & 0.8843 & 0.8786 & 0.8837 & 0.8913 & 0.9123 & 0.9020 & 0.8981 & 0.9041 & 0.9068 & 201 & 210 & 126 & 179.0 \\[-1.5pt]%
 &  &  &  &  &  &  & \sd{0.0048} & \sd{0.0048} &  &  &  & \sd{0.0073} & \sd{0.0070} &  &  &  &  \\%
\cmidrule(lr){1-18}%
$85{\times}85$ / $5{\times}5$ & 867 & $3{\times}3$ & 6.47 & 0.8818 & 0.8843 & 0.8762 & 0.8808 & 0.8871 & 0.9089 & 0.9061 & 0.8965 & 0.9038 & 0.9049 & 204 & 176 & 103 & 161.0 \\[-1.5pt]%
 &  &  &  &  &  &  & \sd{0.0041} & \sd{0.0051} &  &  &  & \sd{0.0065} & \sd{0.0075} &  &  &  &  \\%
\emphrow  &  & $5{\times}5$ & 7.25 & 0.8951 & 0.8938 & 0.8819 & \textbf{0.8903} & \textbf{0.8977} & 0.9101 & 0.9103 & 0.9015 & \textbf{0.9073} & \textbf{0.9090} & 118 & 173 & 156 & 149.0 \\[-1.5pt]%
\emphrow  &  &  &  &  &  &  & \sd{0.0073} & \sd{0.0056} &  &  &  & \sd{0.0050} & \sd{0.0048} &  &  &  &  \\%
\cmidrule(lr){1-18}%
$27{\times}27$ / $3{\times}3$\,$^\dagger$ & 243 & $3{\times}3$ & 6.46 & 0.8724 & 0.8821 & 0.8588 & 0.8711 & -- & 0.8932 & 0.9074 & 0.8900 & 0.8969 & -- & 102 & 156 & 123 & 127.0 \\[-1.5pt]%
 &  &  &  &  &  &  & \sd{0.0117} &  &  &  &  & \sd{0.0093} &  &  &  &  &  \\%
 &  & $5{\times}5$ & 7.24 & 0.8702 & 0.8778 & 0.8662 & 0.8714 & -- & 0.8997 & 0.9048 & 0.8946 & 0.8997 & -- & 202 & 155 & 100 & 152.3 \\[-1.5pt]%
 &  &  &  &  &  &  & \sd{0.0059} &  &  &  &  & \sd{0.0051} &  &  &  &  &  \\%
\midrule%
\grouprow{18}{\muvitenc}%
$65{\times}65$ / $5{\times}5$ & 507 & $3{\times}3$ & 6.38 & 0.8766 & 0.8903 & 0.8906 & 0.8858 & 0.8921 & 0.8964 & 0.9050 & 0.9095 & 0.9036 & 0.9048 & 106 & 122 & 126 & 118.0 \\[-1.5pt]%
 &  &  &  &  &  &  & \sd{0.0080} & \sd{0.0073} &  &  &  & \sd{0.0067} & \sd{0.0055} &  &  &  &  \\%
 &  & $5{\times}5$ & 7.17 & 0.8874 & 0.8911 & 0.8866 & \underline{0.8884} & \underline{0.8946} & 0.9022 & 0.9092 & 0.9041 & \underline{0.9051} & \underline{0.9063} & 105 & 124 & 72 & 100.3 \\[-1.5pt]%
 &  &  &  &  &  &  & \sd{0.0024} & \sd{0.0039} &  &  &  & \sd{0.0036} & \sd{0.0046} &  &  &  &  \\%
\cmidrule(lr){1-18}%
$85{\times}85$ / $5{\times}5$ & 867 & $3{\times}3$ & 6.38 & 0.8786 & 0.8878 & 0.8668 & 0.8777 & 0.8842 & 0.8961 & 0.9066 & 0.8875 & 0.8967 & 0.8985 & 91 & 138 & 113 & 114.0 \\[-1.5pt]%
 &  &  &  &  &  &  & \sd{0.0105} & \sd{0.0109} &  &  &  & \sd{0.0096} & \sd{0.0103} &  &  &  &  \\%
 &  & $5{\times}5$ & 7.17 & 0.8853 & 0.8870 & 0.8852 & 0.8858 & 0.8936 & 0.9058 & 0.9025 & 0.9020 & 0.9034 & 0.9055 & 110 & 80 & 99 & 96.3 \\[-1.5pt]%
 &  &  &  &  &  &  & \sd{0.0010} & \sd{0.0010} &  &  &  & \sd{0.0020} & \sd{0.0028} &  &  &  &  \\%
\cmidrule(lr){1-18}%
$27{\times}27$ / $3{\times}3$\,$^\dagger$ & 243 & $3{\times}3$ & 5.59 & 0.8778 & 0.8744 & 0.8766 & 0.8763 & -- & 0.9003 & 0.9018 & 0.9025 & 0.9015 & -- & 102 & 103 & 88 & 97.7 \\[-1.5pt]%
 &  &  &  &  &  &  & \sd{0.0017} &  &  &  &  & \sd{0.0011} &  &  &  &  &  \\%
 &  & $5{\times}5$ & 6.38 & 0.8891 & 0.8748 & 0.8885 & 0.8841 & -- & 0.9087 & 0.9040 & 0.9035 & 0.9054 & -- & 96 & 103 & 88 & 95.7 \\[-1.5pt]%
 &  &  &  &  &  &  & \sd{0.0081} &  &  &  &  & \sd{0.0029} &  &  &  &  &  \\%
\bottomrule%
\end{tabular}}
\fitwidth{\textwidth}{36}{\cfgtab}
\end{tablewide}

%% file: fig/tab_slice577.tex
\begin{tablewide}[!t]
\centering
\caption{\textbf{The Eight Configurations on a Second Slice of the Held-Out Cell.} Avg.\ Dice as the mean over the three training seeds, with the sample standard deviation (SD) below it, scored on slice~577 and on slice~627 of high\_c4 (the latter as in Table~\mainref{tab:config_selection}). $\Delta$ is the score on slice~577 minus the score on slice~627. Raw is a single forward pass without post-processing, Post is post-processing without TTA3, and TTA3 is test-time augmentation without post-processing. Shading marks the final configuration. Slice~577 played no part in selecting the configuration or in tuning the post-processing.}
\label{tab:slice577}
\tabstyledbl
\newcommand{\slicetab}{%
\renewcommand{\arraystretch}{1.0}%

\begin{tabular}{ll >{\hspace{6pt}}ccc >{\hspace{6pt}}ccc >{\hspace{6pt}}ccc}
\toprule
 & & \multicolumn{3}{c}{Raw} & \multicolumn{3}{c}{Post} & \multicolumn{3}{c}{TTA3} \\
\cmidrule(lr){3-5}\cmidrule(lr){6-8}\cmidrule(lr){9-11}
\makecell[l]{View / patch\\[-1pt]size (px)} & \makecell{S.H.\\[-1pt]kernel} & 577 & 627 & $\Delta$ & 577 & 627 & $\Delta$ & 577 & 627 & $\Delta$ \\
\midrule
\grouprow{11}{MuViT}
\rule{0pt}{9.0pt}$65{\times}65$ / $5{\times}5$ & $3{\times}3$ & 0.8804 & 0.8819 & \dn{0.0016} & 0.9046 & 0.9056 & \dn{0.0010} & 0.8878 & 0.8889 & \dn{0.0011} \\[-1.5pt]
\rule[-3.5pt]{0pt}{0pt} & & \sd{0.0065} & \sd{0.0081} &  & \sd{0.0053} & \sd{0.0060} &  & \sd{0.0066} & \sd{0.0094} &  \\
\rule{0pt}{9.0pt} & $5{\times}5$ & 0.8827 & 0.8837 & \dn{0.0010} & 0.9038 & 0.9041 & \dn{0.0004} & 0.8907 & 0.8913 & \dn{0.0005} \\[-1.5pt]
\rule[-3.5pt]{0pt}{0pt} & & \sd{0.0041} & \sd{0.0048} &  & \sd{0.0087} & \sd{0.0073} &  & \sd{0.0039} & \sd{0.0048} &  \\
\cmidrule{1-11}
\rule{0pt}{9.0pt}$85{\times}85$ / $5{\times}5$ & $3{\times}3$ & 0.8811 & 0.8808 & $+0.0003$ & 0.9057 & 0.9038 & $+0.0018$ & 0.8883 & 0.8871 & $+0.0012$ \\[-1.5pt]
\rule[-3.5pt]{0pt}{0pt} & & \sd{0.0027} & \sd{0.0041} &  & \sd{0.0060} & \sd{0.0065} &  & \sd{0.0039} & \sd{0.0051} &  \\
\emphrow \rule{0pt}{9.0pt} & $5{\times}5$ & 0.8910 & 0.8903 & $+0.0007$ & 0.9080 & 0.9073 & $+0.0007$ & 0.8980 & 0.8977 & $+0.0003$ \\[-1.5pt]
\emphrow \rule[-3.5pt]{0pt}{0pt} & & \sd{0.0064} & \sd{0.0073} &  & \sd{0.0053} & \sd{0.0050} &  & \sd{0.0055} & \sd{0.0056} &  \\
\midrule
\grouprow{11}{\muvitenc}
\rule{0pt}{9.0pt}$65{\times}65$ / $5{\times}5$ & $3{\times}3$ & 0.8846 & 0.8858 & \dn{0.0012} & 0.9036 & 0.9036 & \dn{0.0000} & 0.8912 & 0.8921 & \dn{0.0009} \\[-1.5pt]
\rule[-3.5pt]{0pt}{0pt} & & \sd{0.0095} & \sd{0.0080} &  & \sd{0.0068} & \sd{0.0067} &  & \sd{0.0085} & \sd{0.0073} &  \\
\rule{0pt}{9.0pt} & $5{\times}5$ & 0.8916 & 0.8884 & $+0.0032$ & 0.9091 & 0.9051 & $+0.0040$ & 0.8979 & 0.8946 & $+0.0033$ \\[-1.5pt]
\rule[-3.5pt]{0pt}{0pt} & & \sd{0.0030} & \sd{0.0024} &  & \sd{0.0019} & \sd{0.0036} &  & \sd{0.0031} & \sd{0.0039} &  \\
\cmidrule{1-11}
\rule{0pt}{9.0pt}$85{\times}85$ / $5{\times}5$ & $3{\times}3$ & 0.8766 & 0.8777 & \dn{0.0011} & 0.8994 & 0.8967 & $+0.0026$ & 0.8830 & 0.8842 & \dn{0.0012} \\[-1.5pt]
\rule[-3.5pt]{0pt}{0pt} & & \sd{0.0090} & \sd{0.0105} &  & \sd{0.0077} & \sd{0.0096} &  & \sd{0.0093} & \sd{0.0109} &  \\
\rule{0pt}{9.0pt} & $5{\times}5$ & 0.8872 & 0.8858 & $+0.0014$ & 0.9054 & 0.9034 & $+0.0020$ & 0.8951 & 0.8936 & $+0.0015$ \\[-1.5pt]
\rule[-3.5pt]{0pt}{0pt} & & \sd{0.0023} & \sd{0.0010} &  & \sd{0.0006} & \sd{0.0020} &  & \sd{0.0011} & \sd{0.0010} &  \\
\bottomrule
\end{tabular}}
\sbox0{\setlength{\tabcolsep}{0pt}\slicetab}
\setlength{\tabcolsep}{\dimexpr(\textwidth-\wd0)/22\relax}
\slicetab
\end{tablewide}

%% file: fig/tab_volume.tex
\begin{tablewide}[!t]
\centering
\caption{\textbf{The Final Model on 17 Slices of the Held-Out Cell.} Dice per class and Avg.\ Dice of seed~74 of the final configuration on every 25th slice of high\_c4 from 251 to 651, the range that contains the nucleus class. Slices 577 and 627 are not among them (see \cref{tab:slice577} and Table~\mainref{tab:config_selection}); rows 576 and 626 lie next to them. Raw is a single forward pass without post-processing; the other two blocks use TTA3, the right one with post-processing. Slice numbers count from 1, as in the main paper. Mean and SD are the mean and sample standard deviation over the 17 slices; the SD is over slices, not over seeds, and neighboring slices are correlated. Nuc.\ is nucleus, Gran.\ granule, Mito.\ mitochondria and Unc.\ uncategorized. No slice played a part in selecting the configuration or in tuning the post-processing.}
\label{tab:volume}
\tabstyledbl
\newcommand{\voltab}{%
\renewcommand{\arraystretch}{1.0}%
\begin{tabular}{l >{\hspace{6pt}}ccccc >{\hspace{6pt}}ccccc >{\hspace{6pt}}ccccc}
\toprule
 & \multicolumn{5}{c}{Raw} & \multicolumn{5}{c}{TTA3, no post-processing} & \multicolumn{5}{c}{TTA3 and post-processing} \\
\cmidrule(lr){2-6}\cmidrule(lr){7-11}\cmidrule(lr){12-16}
Slice & Nuc. & Gran. & Mito. & Unc. & \makecell{Avg.\\[-1pt]Dice} & Nuc. & Gran. & Mito. & Unc. & \makecell{Avg.\\[-1pt]Dice} & Nuc. & Gran. & Mito. & Unc. & \makecell{Avg.\\[-1pt]Dice} \\
\midrule
251 & 0.9369 & 0.8245 & 0.8015 & 0.8959 & 0.8647 & 0.9435 & 0.8367 & 0.8196 & 0.9041 & 0.8760 & 0.9496 & 0.8501 & 0.8766 & 0.9194 & 0.8989 \\
276 & 0.9695 & 0.7700 & 0.8186 & 0.8946 & 0.8632 & 0.9711 & 0.7821 & 0.8316 & 0.9013 & 0.8715 & 0.9746 & 0.8026 & 0.8770 & 0.9184 & 0.8931 \\
301 & 0.9652 & 0.7619 & 0.8392 & 0.8875 & 0.8635 & 0.9668 & 0.7729 & 0.8529 & 0.8943 & 0.8717 & 0.9706 & 0.7885 & 0.8808 & 0.9063 & 0.8865 \\
326 & 0.9593 & 0.7885 & 0.8635 & 0.8881 & 0.8748 & 0.9600 & 0.7981 & 0.8752 & 0.8937 & 0.8818 & 0.9616 & 0.8055 & 0.9012 & 0.9037 & 0.8930 \\
351 & 0.9699 & 0.7772 & 0.7808 & 0.8867 & 0.8537 & 0.9704 & 0.7875 & 0.7895 & 0.8928 & 0.8600 & 0.9700 & 0.7953 & 0.8495 & 0.9054 & 0.8801 \\
376 & 0.9817 & 0.7761 & 0.8527 & 0.8873 & 0.8744 & 0.9816 & 0.7854 & 0.8641 & 0.8929 & 0.8810 & 0.9821 & 0.8034 & 0.8881 & 0.9045 & 0.8945 \\
401 & 0.9808 & 0.8151 & 0.8956 & 0.8971 & 0.8971 & 0.9816 & 0.8269 & 0.9027 & 0.9046 & 0.9039 & 0.9866 & 0.8412 & 0.9228 & 0.9160 & 0.9166 \\
426 & 0.9855 & 0.7933 & 0.9016 & 0.8896 & 0.8925 & 0.9854 & 0.8055 & 0.9081 & 0.8962 & 0.8988 & 0.9870 & 0.8128 & 0.9248 & 0.9041 & 0.9072 \\
451 & 0.9794 & 0.8025 & 0.8863 & 0.8859 & 0.8885 & 0.9794 & 0.8144 & 0.8929 & 0.8917 & 0.8946 & 0.9811 & 0.8301 & 0.9018 & 0.9026 & 0.9039 \\
476 & 0.9850 & 0.8246 & 0.9119 & 0.9064 & 0.9070 & 0.9853 & 0.8354 & 0.9175 & 0.9124 & 0.9126 & 0.9867 & 0.8494 & 0.9284 & 0.9230 & 0.9219 \\
501 & 0.9837 & 0.7961 & 0.8990 & 0.8970 & 0.8940 & 0.9844 & 0.8078 & 0.9019 & 0.9023 & 0.8991 & 0.9853 & 0.8191 & 0.9131 & 0.9106 & 0.9070 \\
526 & 0.9800 & 0.7825 & 0.8960 & 0.9018 & 0.8901 & 0.9802 & 0.7946 & 0.9040 & 0.9079 & 0.8967 & 0.9818 & 0.8113 & 0.9181 & 0.9162 & 0.9069 \\
551 & 0.9810 & 0.8022 & 0.9057 & 0.9006 & 0.8974 & 0.9810 & 0.8163 & 0.9129 & 0.9073 & 0.9044 & 0.9839 & 0.8342 & 0.9256 & 0.9193 & 0.9157 \\
576 & 0.9788 & 0.7727 & 0.8987 & 0.9070 & 0.8893 & 0.9793 & 0.7870 & 0.9056 & 0.9134 & 0.8963 & 0.9807 & 0.8029 & 0.9275 & 0.9250 & 0.9090 \\
601 & 0.9729 & 0.7899 & 0.9039 & 0.9019 & 0.8921 & 0.9732 & 0.8055 & 0.9118 & 0.9090 & 0.8999 & 0.9753 & 0.8228 & 0.9219 & 0.9192 & 0.9098 \\
626 & 0.9750 & 0.7954 & 0.9036 & 0.9104 & 0.8961 & 0.9769 & 0.8073 & 0.9102 & 0.9170 & 0.9029 & 0.9796 & 0.8292 & 0.9280 & 0.9296 & 0.9166 \\
651 & 0.9675 & 0.8001 & 0.8622 & 0.9117 & 0.8854 & 0.9695 & 0.8093 & 0.8709 & 0.9173 & 0.8917 & 0.9701 & 0.8252 & 0.8900 & 0.9271 & 0.9031 \\
\midrule
\emphrow Mean & 0.9737 & 0.7925 & 0.8718 & 0.8970 & \textbf{0.8838} & 0.9747 & 0.8043 & 0.8807 & 0.9034 & \textbf{0.8908} & 0.9769 & 0.8190 & 0.9044 & 0.9147 & \textbf{0.9038} \\
SD & 0.0121 & 0.0182 & 0.0404 & 0.0087 & 0.0152 & 0.0107 & 0.0183 & 0.0378 & 0.0087 & 0.0145 & 0.0101 & 0.0184 & 0.0237 & 0.0090 & 0.0114 \\
\end{tabular}}
\sbox0{\setlength{\tabcolsep}{0pt}\voltab}
\setlength{\tabcolsep}{\dimexpr(\textwidth-\wd0)/32\relax}
\voltab
\end{tablewide}